**Physics-informed neural network for predicting fatigue life of unirradiated and irradiated austenitic and ferritic/martensitic steels under reactor-relevant conditions**

Dhiraj S. Kori[1], Abhinav Chandraker[1], Syed Abdur Rahman[2], Punit Rathore[3], and Ankur Chauhan[1]

[1]Extreme Environments Materials Group (EEMG), Department of Materials Engineering, Indian Institute of Science (IISc), Bengaluru, 560012 Karnataka, India

[2]Department of Materials, University of Oxford, Parks Road, Oxford, OX1 3PH, United Kingdom

[3]Robert Bosch Centre for Cyberphysical Systems, Indian Institute of Science (IISc), Bengaluru, 560012 Karnataka, India

*Corresponding author: ankurchauhan@iisc.ac.in

**Abstract**

This study proposes a Physics-Informed Neural Network (PINN) framework to predict the low-cycle fatigue (LCF) life of irradiated austenitic and ferritic/martensitic (F/M) steels used in nuclear reactors. These materials undergo cyclic loading, neutron irradiation, and elevated temperatures, leading to complex degradation mechanisms that are difficult to capture with conventional empirical or purely data-driven models. The proposed PINN embeds fatigue-life governing physical constraints into the loss function, enabling physically consistent learning while improving predictive accuracy, reliability, and generalizability. The model was trained on 495 strain-controlled fatigue data points spanning irradiated and unirradiated conditions. Compared with traditional machine learning approaches, including Random Forest, Gradient Boosting, eXtreme Gradient Boosting, and conventional neural networks, the PINN demonstrated superior performance. SHapley Additive exPlanations (SHAP) analysis identified strain amplitude, irradiation dose, and test temperature as the dominant features, each exhibiting physically meaningful inverse correlations with fatigue life. Univariate and multivariate analyses revealed clear alloy-specific degradation characteristics. Austenitic steels exhibited strong nonlinear coupling among strain amplitude, irradiation dose, and temperature, resulting in pronounced fatigue degradation under combined loading. In contrast, F/M steels demonstrated comparatively stable irradiation responses, including dose-saturation behavior, but showed sensitivity to elevated temperatures beyond tempering thresholds. Despite these advances, predictive reliability is influenced by limited and non-uniform data distributions, as well as the absence of quantitative microstructural descriptors, which are seldom reported in the literature. Although physics-informed constraints improve generalization, predictions under extreme irradiation doses and elevated temperatures should be interpreted with caution. Overall, the proposed PINN framework serves as a reliable and interpretable tool for reactor-relevant fatigue assessment, enabling performance evaluation for advanced nuclear applications, while underscoring the need for expanded datasets and deeper mechanistic integration to further enhance predictive robustness and fidelity.

**Keywords:** Irradiated stainless steels; Fatigue life prediction; Machine learning; Physical constraints; Neural networks

## 1. Introduction

As nuclear energy gains traction as a low-emission power source, preserving the integrity of fusion and fission reactor components under extreme temperature, corrosive coolants, and intense irradiation is essential. Irradiation displaces atoms out of their lattice positions, creating point defects that coalesce into dislocation loops, voids, solute segregation, and irradiation-induced precipitates [1]. These microstructural alterations drive irradiation hardening, embrittlement, and swelling, while also degrading materials' creep and fatigue resistance [2].

To mitigate these effects, structural materials such as austenitic and ferritic/martensitic (F/M) steels are widely used in nuclear reactors due to their complementary properties. Austenitic steels offer high corrosion resistance, efficient heat transfer, and creep resistance, thanks to their high Cr-content and face-centered cubic (FCC) structure. However, their relatively high thermal expansion can impair fatigue performance [3]. In contrast, F/M steels, with a body-centered cubic (BCC) lattice, offer low thermal expansion, superior fatigue resistance, and robust swelling resistance under irradiation, albeit at the expense of some creep strength. These traits make F/M steels especially attractive for next-generation reactor concepts [4].

Despite their advantageous properties, reactor-grade steels still face significant fatigue risks from operational loads, steep thermal gradients, and vibrational stresses [3]. Fusion facilities, such as ITER and DEMO, operate inherently in cyclic modes and contend with plasma instabilities, including Multifaceted Asymmetric Radiation from the Edges (MARFEs), which exacerbate both mechanical and thermal cycling [5,6]. Fission reactor cores likewise undergo repetitive thermal and irradiation-driven stresses, which accumulate as thermomechanical fatigue and ultimately dictate the service life of their structural components [7].

Given this complex interplay of factors, accurately predicting fatigue life under such harsh conditions remains a critical yet challenging task. It demands a deep understanding of material behavior under irradiation, robust modeling frameworks, and advanced characterization techniques to capture the evolving microstructure and its impact on mechanical performance. Traditional numerical approaches, such as those based on Chaboche's viscoplasticity model and continuum damage mechanics, have been applied to both unirradiated [8–10] and irradiated [11] steels. While effective, these models are computationally intensive and require extensive parameter calibration, which depends on multiple low-cycle fatigue (LCF) datasets across temperatures, a major limitation when working with irradiated specimens. An alternative strategy involves estimating parameters for the Manson, Coffin, and Basquin empirical equations [12–14] using tensile properties. This method has been applied to predict the fatigue life of unirradiated F/M steels in both air and vacuum [15–17]. However, these empirical models also rely on costly, large-scale testing and often fall short in capturing the complex interplay of variables such as alloy composition, temperature, coolant environment, and irradiation effects.

In the era of big data and high-performance computing, machine learning (ML) employs advanced algorithms to construct predictive models and rigorously quantify the uncertainties in their forecasts [18–20]. By efficiently mapping complex input features to target outputs, ML offers a compelling alternative to traditional fatigue-life models [21–27] and has proven effective in predicting various forms of irradiation-induced material degradation [28]. For instance, different ML algorithms have been employed to model irradiation hardening [29,30], ductile-to-brittle transition temperature [31] [32], void swelling [33], and irradiation creep [34] in reactor-grade steels. However, the use of ML to predict the fatigue life of irradiated steels is very limited. Zahran et al. [35,36] recently implemented a data-driven ML framework, using Random Forest, Support Vector Regressor (SVR), Gradient Boosted Regressor (GBR), and Multi-Layer Perceptron (MLP), to forecast the LCF life of both unirradiated and neutron-irradiated reduced-activation F/M steels.

Applications to other structural alloys, notably irradiated austenitic steels, remain scarce due to the high cost and lengthy lead times associated with irradiation fatigue testing [37]. Additionally, conventional ML models trained on limited data exhibit poor generalization beyond the training domain. To overcome this, Physics-informed Neural Networks (PINNs) embed physical knowledge into the model, thereby enhancing accuracy, convergence, and robustness, especially in cases with scarce data [38] [39,40]. For instance, Zhang et al. [41] predicted the creep-fatigue life of 316 stainless steel using PINNs by integrating physical constraints into a multi-layer nested neural network, while Chen et al. [42] incorporated probabilistic fatigue behaviour for better interpretability of S-N curves. Advances include the PINN framework by Zhou et al. [43] which integrates fatigue life scatter and physical constraints into the model training process. Their model outperformed purely data-driven ML approaches by preserving physical consistency and improving generalizability. Yang et al. [43] compensated for the limited data sample size by imposing two physical constraints to restrict the model parameter space and by generating domain-knowledge-based data samples. Fan et al. [44] combined the Z-parameter model for fatigue life prediction with an artificial neural network, accounting for stress level, defect size, and location. Fracture mechanics-based constraints were introduced into fatigue life prediction of additively manufactured materials by Salati et al. [45], and sensitive features were combined with multiaxial fatigue constraints within generative adversarial networks by He et al. [45].

Building on these advances, we develop a PINN model to predict the LCF life of unirradiated and irradiated austenitic and F/M steels. By incorporating physical constraints, such as fatigue-related partial derivatives with respect to these variables, into the loss function, the model produces physically consistent and reliable predictions. These predictions are discussed and compared with those obtained using four conventional data-driven ML models: Random Forest (RF), Gradient Boost (GB), eXtreme Gradient Boost (XGB), and a conventional Neural Network (NN).

## 2. Methodology

### 2.1. Input features and data splitting

Fully reversed ($R = -1$), strain-controlled, tension-compression LCF test data for various austenitic steels (AISI SS310, SS304, SS304L, SS310, SS316, SS316L) and F/M steels (HT9, T91, F82H, OPTIFER, and EUROFER97) were compiled from the literature [46–57]. The data comprises 495 data points, including 254 unirradiated and 241 irradiated specimens.

For model development, the data was split into training and testing subsets. The training allowed the ML models to learn underlying patterns, while the testing evaluated their predictive performance in an unbiased manner. Model performance can vary across data splits. To ensure robustness, the "random_state" parameter was varied, and the model was trained and tested across six combinations (two train-test ratios: 80:20 and 70:30, each with three random states). This confirmed the model's generalizability beyond a single split.

The model was trained on over 50 input features (see Table S.1 in the Supplementary Materials), including the elemental composition of the steels in weight percent (wt.%), pre-irradiation treatment parameters such as temperature (°C), time (min), for processing conditions such as cold-working, hot rolling, cold-swaging, normalizing, annealing, tempering, irradiation parameters including radiation dose (dpa), irradiation temperature (°C) and irradiation environments such as Helium, Neon, Argon, Static/flowing Sodium and the absence of an environment/air; testing parameters such as strain amplitude (%), test temperature (°C), strain rate ($s^{-1}$); and specimen type (cylindrical or hourglass) and their respective dimensions (mm). Most of the utilized irradiated data is collected in Static Sodium Environment (39%), followed by the absence of an environment/air (25%), Argon (20%), flowing Sodium (10%), and lastly by Helium-Neon (6%). The range of values for the input features is provided in Table S.2 in the Supplementary Materials. The correlation between experimental input features such as strain amplitude, test temperature, and irradiation dose and fatigue life is provided in Section S.2 in the Supplementary Materials.

Categorical or non-numeric input features, such as irradiation environment and pre-treatment, were one-hot encoded, converting each category into binary variables (1 for presence, 0 for absence) to capture their influence [58]. Processing treatments with available temperature and time data, i.e., annealing, hot rolling, normalizing, and tempering, were treated as separate features. Others lacking detailed parameters, such as solution annealing, hot-then-cold swaging (HTCS), cold working (CW), chill-swaged tempering (CST), were one-hot encoded due to limited data.

To ensure consistency and enhance the performance of the ML models, all input features were normalized using:

$$x_{norm} = \frac{x - x_{min}}{x_{max} - x_{min}} \tag{1}$$

where x is the input feature, and $x_{min}$ and $x_{max}$ are its minimum and maximum values, respectively. This normalization scales all features to a 0–1 range, thereby reducing dimensional differences and enabling faster and more stable convergence.

Since all features were normalized, missing values (NaN) and infinite values were replaced with $-10$ and $+10$, respectively. Missing values typically arise when samples lack pre-treatment data but still have associated temperature and time information, whereas infinite values occur in cases such as the radius of curvature for cylindrical specimens. Assigning $-10$ enables the algorithm to recognize that the missing entry reflects the non-execution or irrelevance of a particular processing step rather than a simple data gap. These replacement values lie well outside the normalized range, intentionally placing such points beyond the main data distribution to minimize their influence during training while preserving numerical continuity. Although values far from the normalized range can potentially introduce training instability, selecting ±10 provides a practical compromise by introducing only minor perturbations without affecting convergence. In contrast, much larger substitutions (e.g., ±100) could destabilize the training process and hinder convergence without offering additional benefit. Further details are provided in Section S.3 of the Supplementary Material. For alloy composition features, the absence of a particular element was assigned a value of 0 rather than $-10$, as this directly represents the physical absence of that element in the alloy and therefore conveys meaningful information to the training algorithm.

Radiation dosage (dpa) was used instead of fluence to simplify interpretation and improve accuracy, as neutron fluence spans a wide range up to $1.01 \times 10^{23}$ n/cm² or 71 dpa. A fluence of $10^{22}$ n/cm² corresponds to ~7 dpa for $E \geq 0.1$ MeV [59]. Dosage values were computed accordingly. The output variable, fatigue life (N), was logarithmically scaled to reduce variability and improve prediction accuracy.

Once the input features and data splitting is completed, the ML algorithms are trained and tested. The brief descriptions of the employed ML algorithms including RF, GB, XGB, and NN are provided in detail in Section S.4 of the Supplementary Material.

## 2.2. Physics-Informed Neural Network (PINN) Architecture

Inspired by Raissi et al. [60] and Jiang et al. [61], the PINN architecture (see Fig. 1) was developed by embedding a physics-informed loss function into the neural network. To capture the dependence of fatigue life on different test conditions, partial derivatives with respect to strain amplitude ($\epsilon$), temperature (T), and radiation dose (d) are used. These derivatives are incorporated as inequality constraints ($\frac{\partial \mathrm{N}}{\partial \epsilon} \leq 0$, $\frac{\partial^2 \mathrm{N}}{\partial \epsilon^2} \geq 0$, $\frac{\partial \mathrm{N}}{\partial \mathrm{T}} \leq 0$, $\frac{\partial \mathrm{N}}{\partial \mathrm{d}} \leq 0$) during training to enforce physically consistent learning. These constraints represent established trends: fatigue life decreases with increasing strain amplitude, temperature, or irradiation dose, and increases as the strain–life (S–N) curve becomes steeper. This formulation yields a set of partial differential equations (PDEs) governing the S–N relationship, which can be solved to predict fatigue life under varying conditions.

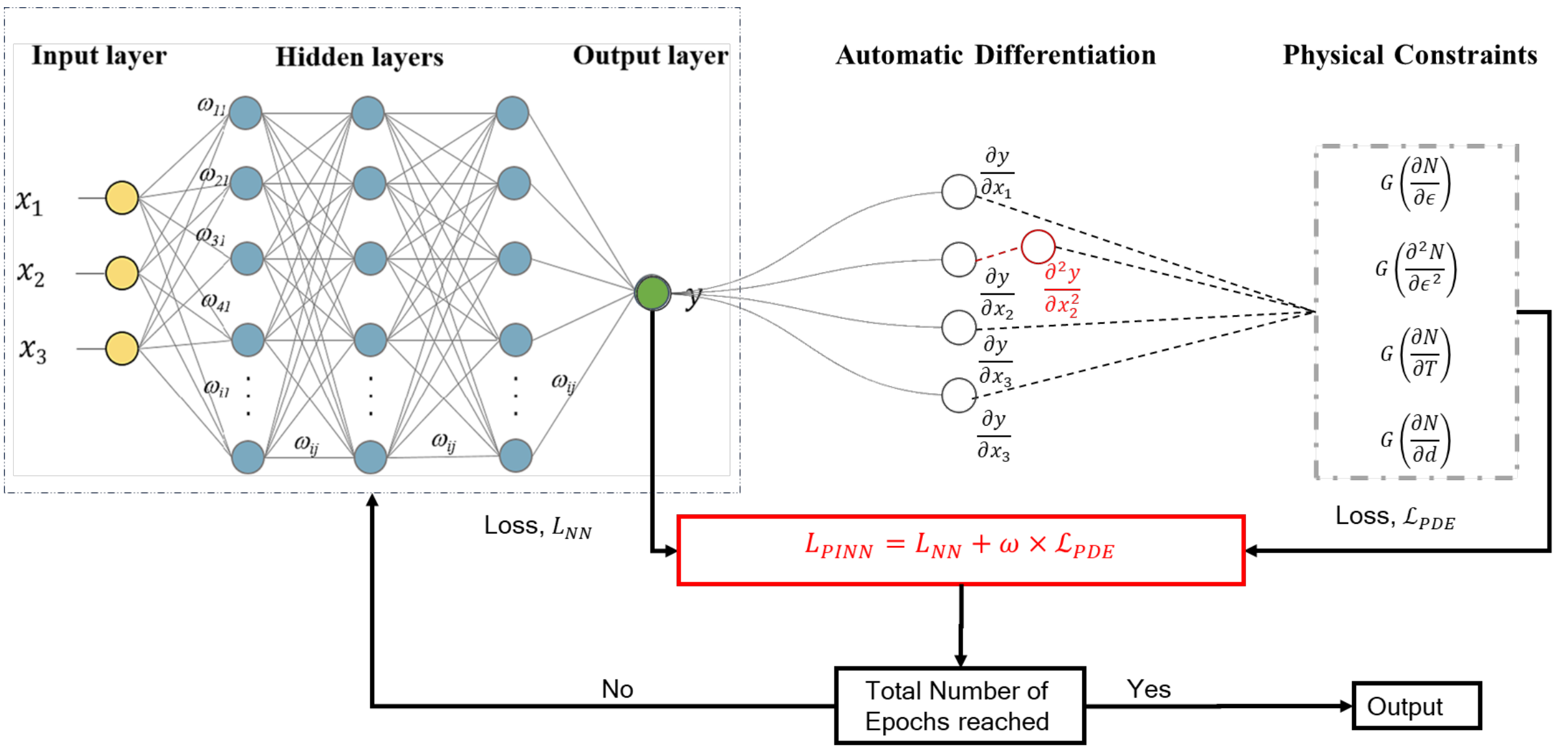

Fig. 1. The architecture of the PINN model consists of input, hidden, and output layers of NN, along with physical constraints.

PyTorch's automatic differentiation is used to compute the PDEs by taking derivatives of the model output with respect to each input feature. A detailed derivation to compute the PDEs is provided in the Supplementary Material Section S.5. The final expression for the first derivative is given as:

$$\nabla_{x_i} y = \prod_{l=k}^{1} \left[ J_{a^{(l)}}\left(z^{(l)}\right) J_{z^{(l)}}\left(a^{(l-1)}\right) \right] \tag{2}$$

Further differentiating the above derivative to get the second derivative:

$$\nabla^2_{x_i} y = \sum_{p=k}^{1} \left[ F_{k,p}\left[H_p(x_i)\right] F_{p-1,1} \right] \tag{3}$$

where,

$$H_p(x_i) = \frac{\partial}{\partial x_i}\left(J_{a^{(p)}}\left(z^{(p)}\right)J_{z^{(p)}}\left(a^{(p-1)}\right)\right) \tag{4}$$

And

$$F_{k,p} = \prod_{l=k}^{p}\left[J_{a^{(l)}}\left(z^{(l)}\right)J_{z^{(l)}}\left(a^{(l-1)}\right)\right] \tag{5}$$

In this framework: The Jacobian matrix, is defined as:

$$J_f(x) = \frac{\partial f}{\partial x} = \left[\frac{\partial f}{\partial x_1}, \frac{\partial f}{\partial x_2}, \cdots\cdots, \frac{\partial f}{\partial x_n}\right] \tag{6}$$

- **Input features** are denoted by $x_i$ for $i^{th}$ input feature and represent the raw variables fed into the network.
- **Pre-activation**, denoted by *z*, refers to the weighted sum of inputs to a neuron—calculated before applying any activation function.
- **Activation**, denoted by *a*, is the neuron's output after applying the activation function to $z$.
- The **final output** of the network is represented by *y*, which corresponds to the network's prediction.

The function, denoted as G(x) in the architecture, transforms the derivatives within the loss function by scaling them to a range between 0 and 1. This non-linear transformation helps bind the derivatives, improving the stability and convergence of the optimization process. It integrates seamlessly into the loss function and is defined as:

$$\mathrm{G(x)} = \frac{\mathrm{x}}{1 + \mathrm{e}^{\beta|\mathrm{x}|}} \tag{7}$$

Where x is the input derivative value, and β controls the shape and steepness of the activation function and the influence of the physics-informed loss during training. For this model, β was set to 100, following Jiang et al. [61], as it yielded optimal results (See Supplementary Materials Section S.6 for more detailed analysis).

To effectively integrate physics-based constraints into neural network training, a custom loss function was developed that combines data-driven and physics-informed components. The total loss function is expressed as:

$$\mathcal{L}(y_{true}, y_{pred}) = H(y_{true}, y_{pred}) + R \tag{8}$$

Here, *H* denotes the Huber loss, which calculates the loss between predicted ($y_{pred}$) and actual ($y_{true}$) fatigue life values [62]. It integrates squared error and absolute error terms to prevent overfitting and improve robustness to outliers. It smoothly transitions between squared and absolute error regimes based on a threshold parameter δ, as defined by:

$$H(y_{true}, y_{pred}) = \begin{cases} \frac{1}{2}(y_{true} - y_{pred})^2, \text{if } |y_{true} - y_{pred}| \leq \delta \\ \delta\left(|y_{true} - y_{pred}| - \frac{\delta}{2}\right), \text{ otherwise} \end{cases} \tag{9}$$

In this study, δ = 1 was selected to define the transition point between the quadratic and linear regimes.

A regularization term, R, in Equation (4) was added to penalize excessive deviation of predictions from their means, thereby further regulating model complexity and improving generalization. This term discourages overfitting to noisy data and smoothens the learned function. It is given by:

$$R = q\left(\frac{1}{n}\sum_{i=1}^{n} y_{pred,\,i}\right)^2 \tag{10}$$

where $y_{pred,\,i}$ is the predicted output for the $i^{th}$ sample, n is the number of samples, and q is a weighting factor (set to 0.01) that controls the influence of the regularization term in the loss function.

The final physics-derived loss, $Loss_{PDE}$, enforces known fatigue behavior by constraining on the partial derivatives of the predicted fatigue life N with respect to the relevant test parameters: ϵ, T, and d. Each derivative is first transformed via a sigmoid-like function G(x) to bound its value between 0 and 1. Then, the physics-informed loss is computed as:

$$\text{Loss}_{\text{PDE}} = \mathcal{L}\left(0,\ G_{\text{out}}\left(\frac{\partial N}{\partial \epsilon}\right)\right) + \mathcal{L}\left(0,\ G_{\text{out}}\left(\frac{\partial^2 N}{\partial \epsilon^2}\right)\right) + \mathcal{L}\left(0, G_{\text{out}}\left(\frac{\partial N}{\partial T}\right)\right) + \mathcal{L}\left(0, G_{\text{out}}\left(\frac{\partial N}{\partial d}\right)\right) \tag{11}$$

where $G_{\text{out}}$ denotes the transformed derivative, $\mathcal{L}(0,\ \ldots.)$ uses the same combined Huber + regularization form from Equation (4), with a zero "true" target to enforce constraints. By construction, these terms drive the network to satisfy: $\frac{\partial N}{\partial \epsilon} \leq\ 0, \frac{\partial^2 N}{\partial \epsilon^2} \geq\ 0, \frac{\partial N}{\partial T} \leq\ 0, \frac{\partial N}{\partial d} \leq\ 0.$ ensuring that fatigue life decreases (or at least does not increase) as strain amplitude, temperature, or dosage increases, and its curvature in strain amplitude is non-negative.

Finally, the total PINN loss is defined by combining the standard neural network loss, $\text{Loss}_{\text{PINN}}$, with the physics-derived penalty term:

$$\text{Loss}_{\text{PINN}} = \text{Loss}_{\text{NN}} + \omega \times \text{Loss}_{\text{PDE}} \tag{12}$$

Here, the hyperparameter $\omega$ determines the relative importance of the physics-informed component: a higher $\omega$ enforces stronger adherence to physical constraints, while a lower $\omega$ allows for more flexibility in fitting the data. By tuning $\omega$, the model strikes a balance between data fidelity and physically consistent predictions.

When the dataset is large and reliable, smaller $\omega$ values generally yield better performance. Conversely, for limited or noisy data, higher $\omega$ values improve model robustness by compensating for uncertainty. To maintain adaptability, $\omega$ was tuned within the range of 0–1, with optimized values across different data splits falling between 0.05 and 0.23.

To enhance training stability and performance, Nakamura et al.'s [63] learning rate schedule was utilized, which combines a warmup phase with a sigmoid decay. In this approach, the learning rate increases linearly at the start (warmup phase) and then gradually decreases following a sigmoid function. This approach improves convergence and test accuracy.

To benchmark the PINN, its performance was compared against four data-driven ML models: RF, GB, XGB, and NN, all trained to predict cycles to failure. Details of these models are provided in Section S.4 of the Supplementary Materials.

Model performance was evaluated using two standard metrics: the coefficient of determination ($R^2$) and Mean Squared Error (MSE). Hyperparameters were optimized via Bayesian search to ensure both accuracy and generalization. The optimized hyperparameters for all models are provided in Section S.7 of the Supplementary Materials (see Tables S.3 to S.7).

The $R^2$ measures how well the predicted fatigue life ($\hat{y}_i$) explains the variance in the experimental/true fatigue life ($y_i$), and is computed as:

$$R^2 = 1 - \frac{\sum_{i=1}^{n}(y_i - \hat{y}_i)^2}{\sum_{i=1}^{n}(y_i - \bar{y}_i)^2} \tag{13}$$

Where, $\bar{y}_i$ is the mean of the experimental/true fatigue life, and $n$ is the number of samples. MSE measures the average squared difference between predicted and actual values, representing overall prediction error:

$$MSE = \frac{1}{n}\sum_{i=1}^{n}(y_i - \hat{y}_i)^2 \tag{14}$$

Additionally, Block-wise cross-validation was performed to evaluate the predictive performance of the proposed PINN model. The dataset was randomly shuffled and divided into five equal blocks, with each block serving as the validation set once. In every iteration, one block was used for validation while the remaining four were used for training. This approach ensures a fair evaluation of the model’s generalisation while minimising bias introduced by data partitioning.

## 2.3. Interpretability and PINN extrapolation capability

To interpret model predictions and feature roles, SHapley Additive exPlanations (SHAP) analysis was utilized. Based on cooperative game theory, SHAP assigns each input feature a contribution value for a given prediction (see Supplementary Materials Section S.4.5 for more details). These are visualized using summary plots, where each horizontal data point spread shows the overall impact of a feature across all samples. It is important to note that the SHAP analysis is not intended to uncover new insights, but to enhance model interpretability and transparency. It provides a critical validation step by confirming that the PINN relies on physically meaningful variables rather than spurious correlations.

A univariate trend analysis assessed the individual effects of features on fatigue life by varying irradiation dosage (dpa) and test temperature (°C) across different strain amplitudes, while keeping other inputs constant. A synthetic dataset within the original training range was generated, and the trained PINN model predicted cycles-to-failure for these inputs.

To better understand the factors governing fatigue behavior, a multivariate trend analysis was also conducted using three key experimental parameters: strain amplitude (%), irradiation dose (dpa), and test temperature (°C). Unlike univariate analyses, which examine each factor separately, this approach captures how these parameters interact to jointly affect fatigue life. A detailed 3D parametric grid was generated over the entire experimental range, and the trained PINN model was used to predict fatigue life at each grid point. To interpret these predictions, SHAP values were calculated across the grid, quantifying the local influence of each parameter and revealing how their relative contributions vary with different combinations of strain, dose, and temperature.

## 3. Results and discussion

### 3.1. PINN training-testing and comparison with other ML models

Model performance was evaluated using an 80:20 training-to-testing data split generated from the representative fourth random state. Fig. 2a-e compares their predictive accuracy on both training and testing datasets. The detailed $R^2$ and MSE values for training and testing for all five models —RF, GB, XGB, NN, and the proposed PINN —are provided in Tables S.8–S.11 of the Supplementary Materials.

Among the models (see Fig. 2a-e), the PINN achieved the best overall performance (Fig. 2e), with a test $R^2$ of 0.88 and an MSE of 0.059, as well as strong training accuracy ($R^2$ = 0.889, MSE = 0.055). In contrast, the RF, GB, XGB, and NN models exhibited slightly lower predictive performance and broader scatter (Figs. 2a–d). Their test $R^2$ values ranged from 0.849 to 0.866, and MSE from 0.066 to 0.075. Training metrics for these models were generally higher, RF ($R^2$ = 0.973, MSE = 0.013), GB ($R^2$ = 0.939, MSE = 0.003), XGB ($R^2$ = 0.995, MSE = 0.003), and NN ($R^2$ = 0.874, MSE = 0.062), indicating slight overfitting, as their test accuracy dropped notably compared to training (See Section S.9 of the Supplementary Materials for a more detailed analysis of the training convergence of the PINN loss components).

The PINN, by contrast, maintained a close match between training and testing results, demonstrating better bias–variance balance. Its tight clustering along the parity line, with nearly 99% of its predictions falling within ±1 standard deviation error band (Fig. 2e), emphasizes its superior ability to capture the underlying physics–data correlations, resulting in more accurate and stable fatigue-life predictions across varying conditions.

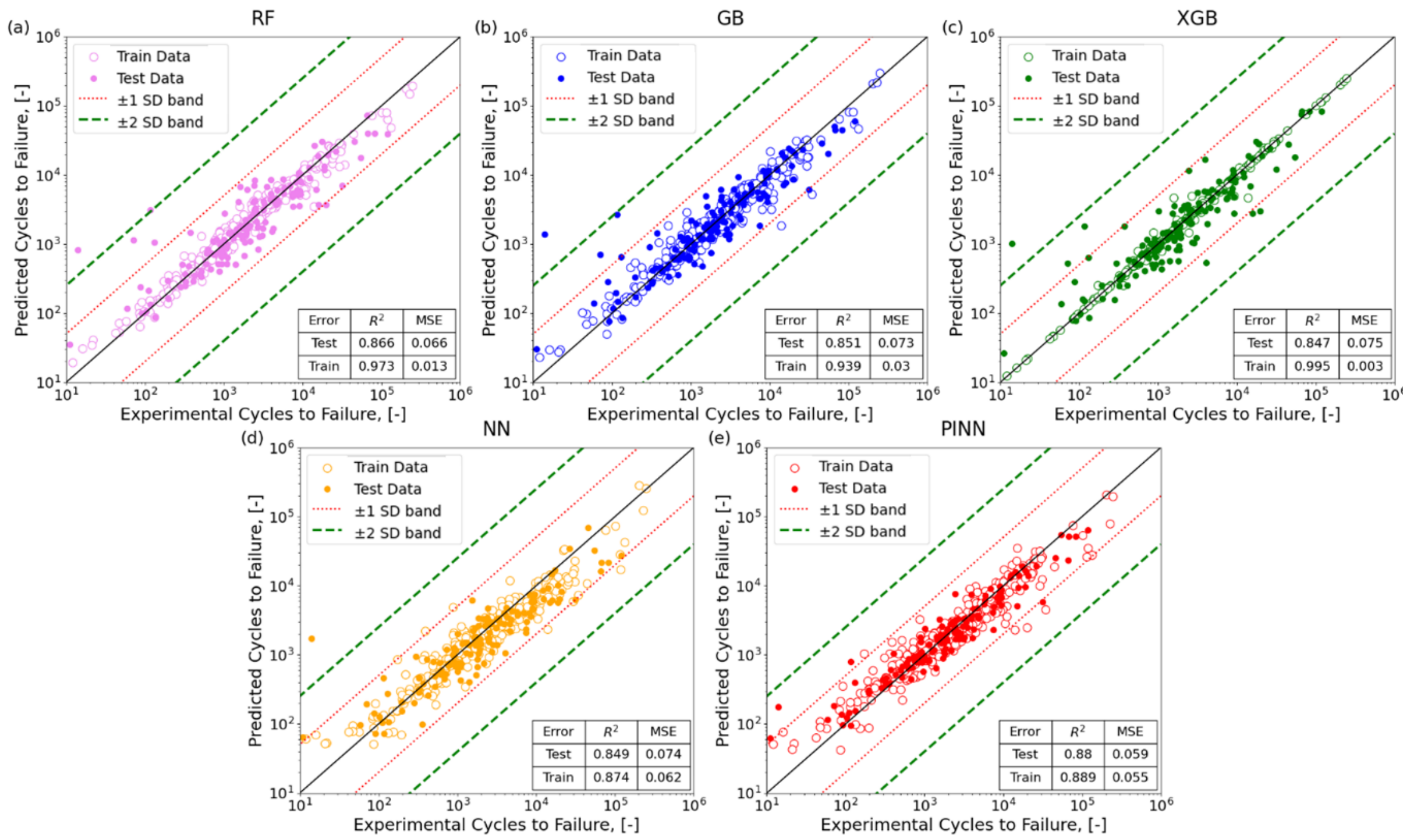

Fig. 2: Comparison between predicted and experimental fatigue life for the testing data using (a) RF, (b) GB, (c) XGB, (d) NN, and (e) PINN models. Red dotted lines indicate the ±1 geometric standard deviation (SD) error band, and green dashed lines denote the ±2 geometric SD band. Results correspond to the 80:20 training–testing split from the representative fourth random state. Nearly all predictions fall within the ±2 standard deviation error band, with the PINN model showing the highest accuracy and consistency among all compared approaches.

To assess model robustness and sensitivity to data partitioning, performance variability across multiple random states was analyzed using the mean and standard deviation of $R^2$ and MSE values, as shown in Fig. 3. Although the PINN achieved strong performance in the representative case above ($R^2$ = 0.88, MSE = 0.059; Fig. 2e), its consistency across different splits was particularly noticeable. Among all models, the PINN maintained the highest average $R^2$ (~0.85) within a narrow range of 0.82–0.88, and the lowest average MSE (~0.07), ranging from 0.046–0.087, reflecting good accuracy and robustness.

In contrast, the data-driven ML models exhibited greater variability. The XGB model showed moderate variance and good generalization ($R^2$: 0.77–0.90; MSE: 0.034–0.111), followed by GB ($R^2$: 0.79–0.885; MSE: 0.039–0.102) and RF ($R^2$: 0.797–0.894; MSE: 0.036–0.099). The NN model displayed the least stability, with $R^2$ ranging from 0.75–0.87 and MSE from 0.043–0.125, indicating higher sensitivity to data partitioning.

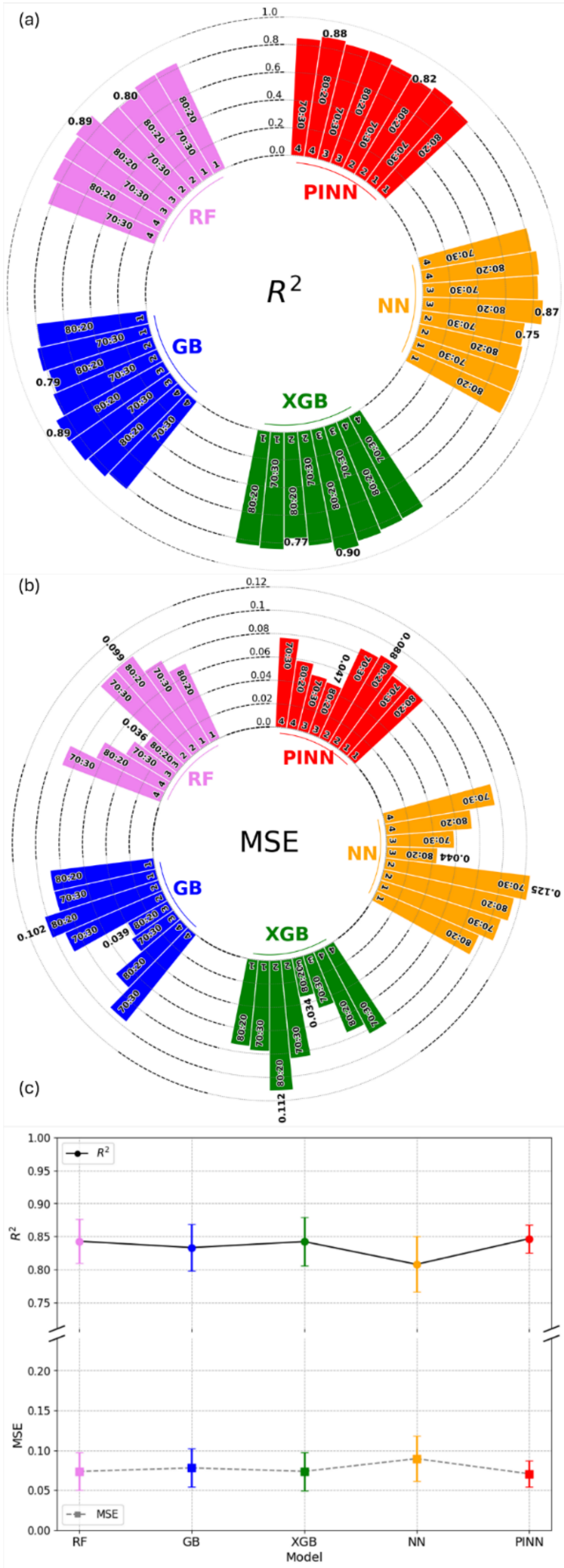

Fig. 3: (a) R² and (b) MSE for five predictive models, RF, GB, XGB, NN, and PINN, are presented. Each of the eight columns corresponds to a distinct combination of two train-test splits (70:30 and 80:20) across four different random states. (c) The average R² and MSE values across these combinations are shown with error bars representing the variance.

The PINN also recorded the lowest standard deviation ($\sigma = 0.020$), outperforming RF ($\sigma = 0.031$), GB ($\sigma = 0.032$), and XGB ($\sigma = 0.035$), indicating greater stability and reduced dependence on specific data splits. This trend was further confirmed by its coefficient of variation (2.36%), which is substantially lower than those of RF (3.56%), GB (3.60%), and XGB (4.16%), demonstrating the superior robustness and reliability of the PINN model.

### 3.2. PINN model validation through block-wise cross-validation

To assess the generalization capability of the proposed PINN model, a block-wise cross-validation strategy was additionally employed. The whole dataset was first randomized and then partitioned into five equal-sized blocks, ensuring uniform representation of strain amplitude and fatigue life across all subsets. In each iteration, the model was trained on four blocks and tested on the remaining one. As shown in Fig. 4a, the predicted fatigue lives from all validation blocks agree well with the experimental data, with most points lying within ±1 standard deviation of the experimental data. The corresponding statistical results in Fig. 4b show that the model maintains stable predictive accuracy across all blocks, achieving $R^2$ values between 0.76 and 0.86 and MSE values between 0.06 and 0.125**.** These relatively narrow ranges indicate consistent model performance without significant degradation across different data partitions.

Furthermore, Fig. 4c illustrates the distribution of experimental fatigue life as a function of strain amplitude, overlaid with predictions from all five validation blocks. The clustering of predicted values closely along the experimental trend confirms that the PINN effectively learns the underlying relationship between strain amplitude and fatigue life. Collectively, these results highlight the model's robustness and strong generalization capability, demonstrating that it performs reliably even when trained and validated on different subsets of data.

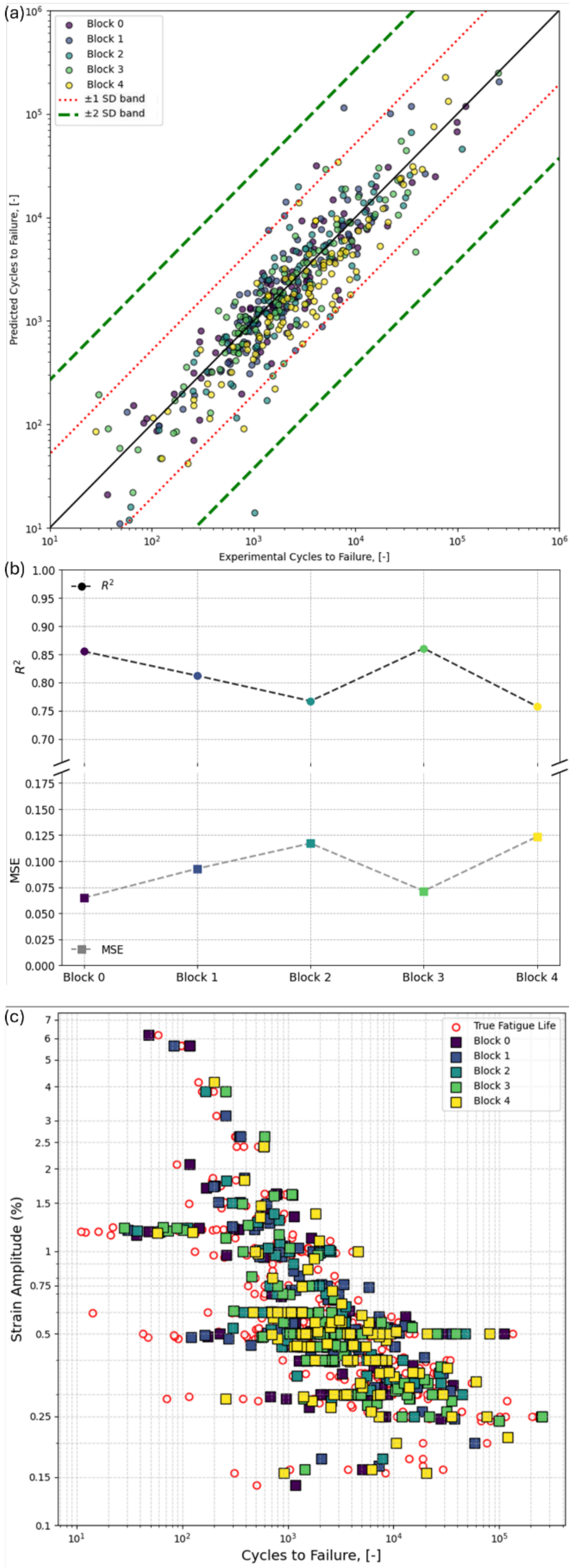

Fig. 4: Block-wise cross-validation results of the proposed PINN model. (a) Parity plot comparing predicted and experimental fatigue lives for all validation blocks, showing that nearly all predictions lie within the ± 1 geometric standard deviation error band, indicating strong predictive agreement. (b) Variation of $R^2$ and MSE values across the five validation blocks, demonstrating consistent model accuracy and robustness to data partitioning. (c) S–N representation of the experimental fatigue life (True Fatigue Life) along with predictions from all validation blocks, highlighting close alignment between predicted and experimental trends across the full strain amplitude range.

### 3.3. SHAP analysis – Input features' importance and their impact

Further insights into the PINN model's interpretability were obtained through SHAP analysis, as illustrated in Fig. 5. The SHAP plot highlights the relative importance and influence of each input feature on the model's output. Features are ranked vertically, with those contributing most to the model's performance appearing at the top. Each horizontal spread of coloured dots represents the impact of that feature across the data. Red dots on the right side of the vertical axis indicate that higher feature values contribute positively to the predicted LCF life, while blue dots suggest a negative correlation. Therefore, this visualisation enables the classification of features into two groups based on their directional impact. One exhibits a positive correlation with the output, where an increase in the input feature value results in an increase in the fatigue lifetime, whereas the other demonstrates a negative correlation with the output.

The SHAP summary plot in Fig. 5 reveals that strain amplitude (%), test temperature (°C), and irradiation dosage (dpa) are the three most influential input features affecting the output, i.e., the predicted LCF life. These features exhibit predominantly negative SHAP values at higher feature magnitudes, indicating an inverse relationship: increases in these parameters are associated with a reduction in the number of cycles to failure, consistent with established fatigue and irradiation physics

Among them, strain amplitude is the most dominant predictor. This also aligns well with the work of Zahran et al [35]. At higher strain amplitudes, materials are subjected to larger cyclic stresses and inelastic strains [64], promoting intensified slip activity and more severe cyclic plastic deformation. This accelerates the development of three-dimensional dislocation substructures, such as dislocation walls, cells, and persistent slip bands, which, upon impinging on the specimen surface, act as preferential sites for crack initiation [65,66], thereby reducing the fatigue life.

Test temperature emerges as the second most influential factor. At elevated temperatures, reductions in shear modulus and increases in atomic diffusion accelerate dislocation mobility via thermally activated processes, such as climb and cross-slip [67,68]. These mechanisms promote dynamic recovery and annihilation of dislocations, accelerating the formation and rearrangement of three-dimensional dislocation structures [67] and the movement of dislocations to the specimen surface, creating preferential sites for crack initiation. In F/M steels, dynamic recovery and annihilation of dislocations at high temperatures also result in undesirable cyclic softening behaviour [69]. This thereby accelerates the nucleation and propagation of cracks. Moreover, environmental effects, such as oxidation, further reduce fatigue resistance at high temperatures compared to specimens tested at ambient conditions [68,70].

The third influential factor, irradiation dosage, exerts a pronounced detrimental effect on fatigue life in the LCF regime. Higher irradiation levels promote the formation and migration of defect clusters, such as interstitial/vacancy complexes, dislocation loops, and cavities [71]. Although these defects generally act as obstacles to mobile dislocations and contribute to irradiation hardening, they simultaneously reduce the material's strain-hardening capacity, leading to lower ductility and promoting early strain localisation [71] by creating features such as dislocation channelling. Given that ductility controls the fatigue life in the LCF regime, reduced ductility directly correlates with shorter fatigue life. Furthermore, irradiation-induced segregation and the accumulation of helium or point defects at grain boundaries can lead to embrittlement and intergranular cracking, exacerbating fatigue damage mechanisms [72] [54].

Other input features appear to have less influence on the output, as indicated by the much narrower spread of the red/blue dots around the vertical partitioning line. However, this could be due to lower variability in the utilised dataset. Additionally, Section S.10 of the Supplementary Materials presents the performance of all models before and after removing highly correlated input features.

Overall, the SHAP analysis confirms that the PINN model captures physically meaningful complex dependencies, reinforcing its predictive accuracy with mechanistic fidelity. The strong alignment between predictions and known deformation–fatigue behavior in irradiated materials underscores the value of integrating domain knowledge into neural network frameworks for reliable and accurate fatigue life estimation.

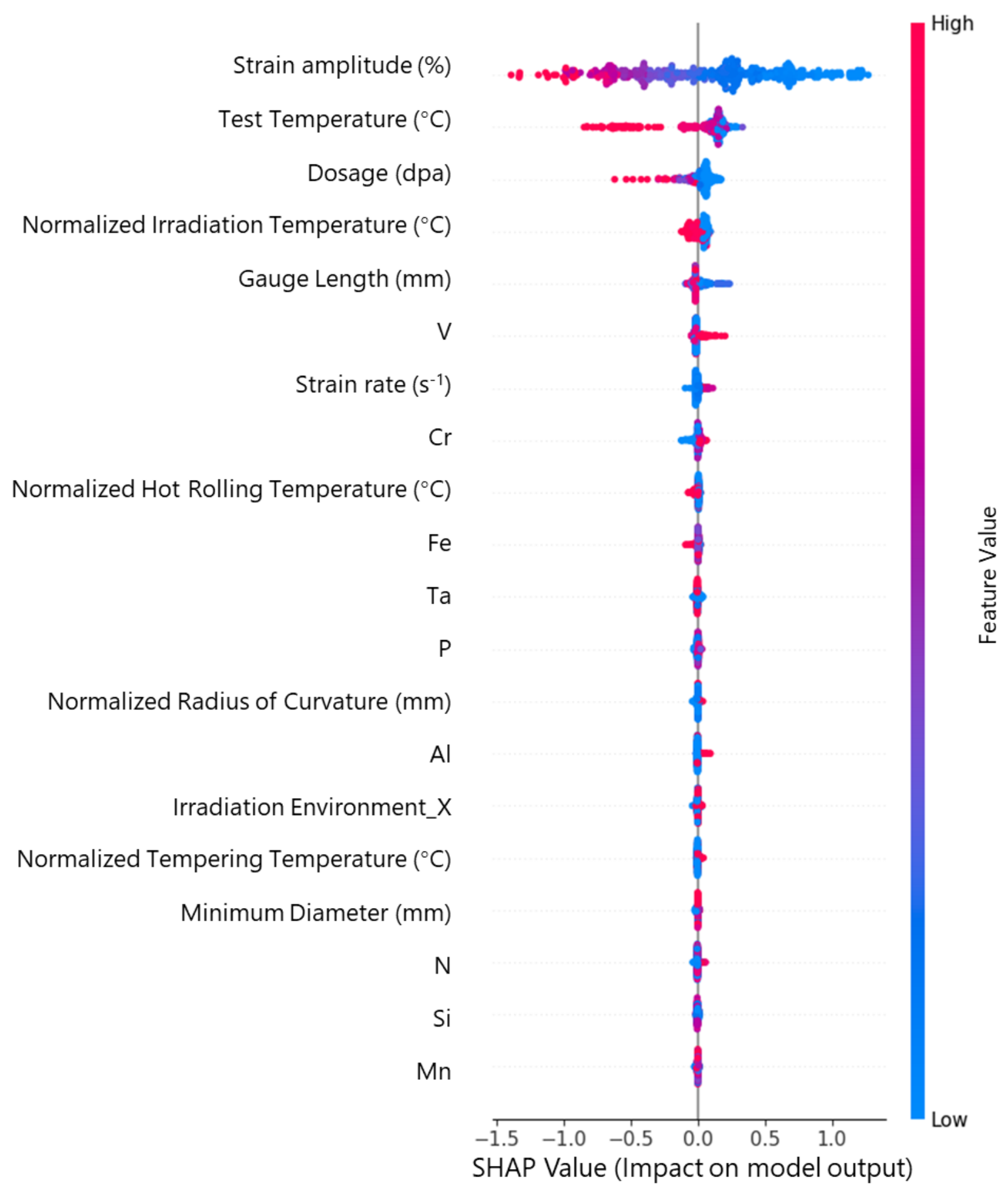

Fig. 5: SHAP summary plot showcasing the 20 topmost influential input features affecting the predicted fatigue life in the PINN model. Each point represents the SHAP value of a single data instance, coloured by the corresponding feature value (red for high, blue for low). Points to the right of the vertical line indicate a positive contribution to the output, while those to the left indicate a negative impact. A greater horizontal spread and a higher ranking of the input feature indicate a significant influence on the model's predictions.

### 3.4. Univariate trends obtained using synthetic data

Fig. 6 presents the predicted fatigue life (N) as a function of strain amplitude (S) for SS316 (austenitic stainless steel) and EUROFER97 (ferritic/martensitic steel), obtained using the developed PINN model. The predictions were generated using univariate trend analysis, in which irradiation dose and test temperature were systematically varied while keeping all other parameters constant.

The analysis incorporates experimentally derived features specific to each material. For SS316, the features correspond to hourglass specimens that were hot-rolled, chilled, swaged, and tempered, then irradiated in a flowing sodium environment at 450 °C and tested at a strain rate of 0.08%/s. In contrast, the features for EUROFER97 correspond to a normalised and tempered cylindrical specimen irradiated in a static sodium environment at 340 °C and tested at a strain rate of 0.003%/s.

Figs. 6a–b show the SS316 predictions: fatigue life versus strain amplitude for doses of 0–70 dpa at 500 °C (Fig. 6a), and versus temperature from 300–700 °C at 30 dpa (Fig. 6b). Figs. 6c–d present the corresponding trends for EUROFER97 under the same dose and temperature ranges.

Evidently, the PINN model accurately reproduces expected fatigue degradation trends with strong physical consistency. Across both alloys, fatigue life decreases with increasing strain amplitude, reflecting the model's ability to capture fundamental cyclic deformation behavior. Moreover, the effects of irradiation dose and temperature are clearly captured, highlighting alloy-specific degradation sensitivities.

For SS316 (Figs. 6a), the predicted fatigue life shows a notable reduction with increasing irradiation dose at a fixed test temperature of 500 °C. This degradation reflects the alloy's susceptibility to irradiation-induced microstructural damage, including defect formation and segregation, which reduce work hardening capability and ductility and induce embrittlement. The adverse effect of test temperature is also evident in Fig. 6b, where increasing the temperature from 300 °C to 700 °C at a fixed dose of 30 dpa further accelerates fatigue damage across all examined strain amplitudes.

EUROFER97 (Figs. 6c–d) exhibits fatigue lives broadly comparable to SS316 at low strain amplitudes but lower at higher strain amplitudes. However, its irradiation response is markedly more stable: fatigue life decreases relatively mildly with increasing dose and temperature, reflecting its superior irradiation tolerance to SS316. Even at doses up to 70 dpa (Fig. 6c), the predicted fatigue life remains close to unirradiated condition. This trend aligns with experimental reports showing that reduced-activation F/M steels retain fatigue lives clustered around those of their unirradiated counterparts [57]. Notably, a slight improvement in fatigue life is observed at higher strain amplitudes under elevated doses (Fig. 6c), consistent with the observations by Materna-Morris et al. [47].

The superior irradiation tolerance of EUROFER97 stems from its BCC crystal structure and hierarchical F/M microstructure, which provides abundant defect sinks, including solutes, dislocation cores, precipitates/matrix interfaces, and various boundaries, that promote point defect recombination [73–75]. These combined effects suppress irradiation-induced damage, including void swelling, thereby preserving mechanical integrity.

Overall, the PINN model accurately captures the univariate effects of irradiation dose, temperature, and strain amplitude on the fatigue life of the two steels. This highlights its potential for reliable fatigue-life assessment under reactor-relevant operating conditions where experimental data are not available.

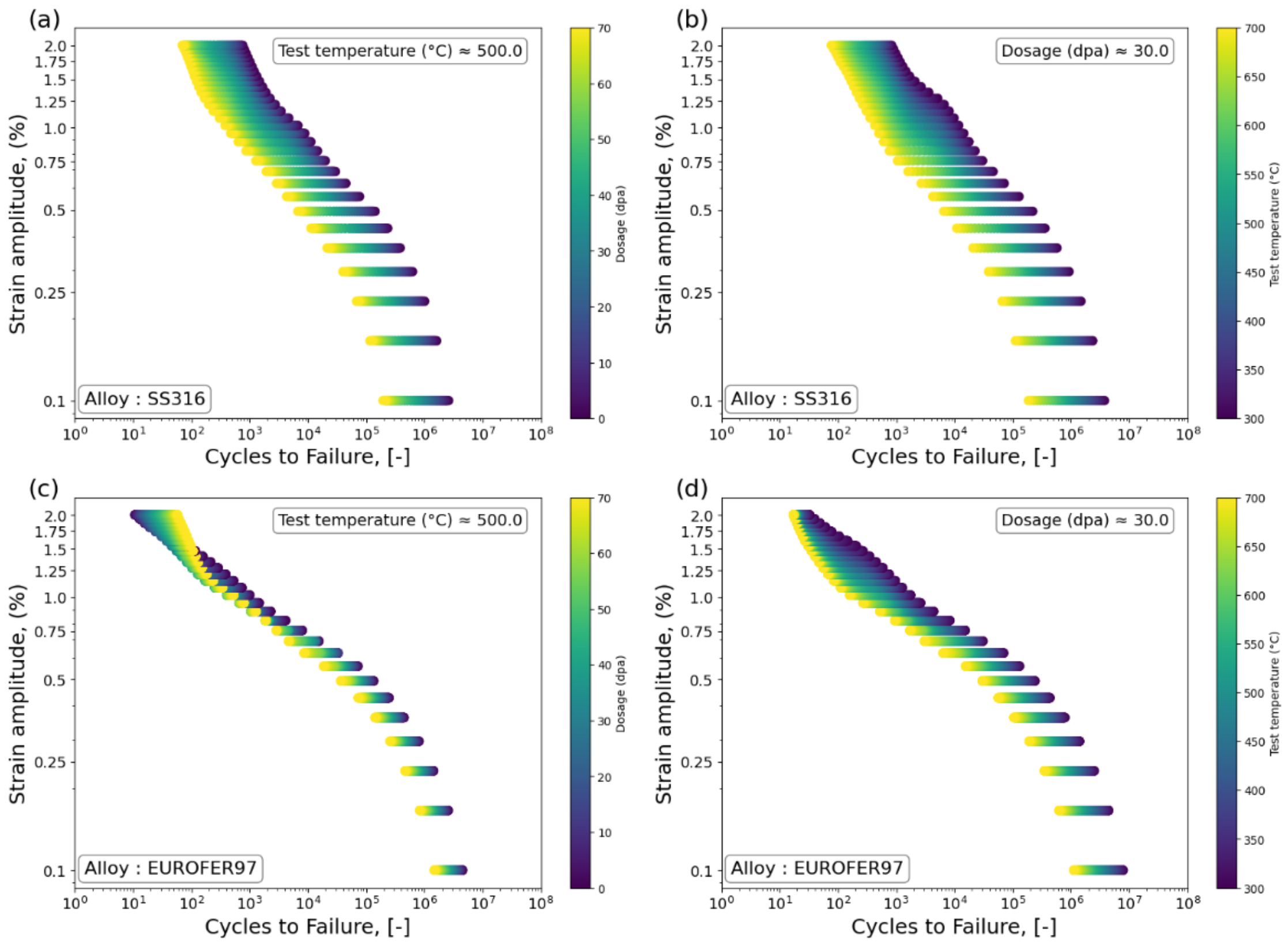

Fig. 6: Predicted S–N curves illustrating the influence of irradiation dose and test temperature on fatigue life for (a, b) austenitic SS316 and (c, d) ferritic/martensitic (F/M) EUROFER97 steels, as obtained from the PINN model. In (a) and (c), the test temperature is fixed at 500 °C while the irradiation dose is varied from 0 to 70 dpa. In (b) and (d), the irradiation dose is held constant at 30 dpa, and the test temperature is varied from 300 to 700°C. Fatigue degradation is more pronounced in SS316, whereas EUROFER97 exhibits higher fatigue resistance, reflecting its superior mechanical integrity under irradiation.

### 3.5. Multivariate feature analysis: Synergistic and Antagonistic effects of input features

Building on the univariate trends presented in Fig. 6, Fig. 7 examines the coupled influence of strain amplitude, irradiation dose, and test temperature on fatigue life through multivariate surface representations. The figure presents three-dimensional PINN predictions for austenitic SS316 (a–c) and ferritic/martensitic EUROFER97 (d–f), where two parameters are varied simultaneously while the third is held constant, enabling assessment of interaction effects beyond univariate behaviour.

Figs. 7a and 7d illustrate the combined effects of strain amplitude and irradiation dose at a fixed test temperature of 500 °C. For both steels, fatigue life decreases sharply with increasing strain amplitude, confirming strain amplitude as the dominant controlling variable. In SS316 (Fig. 7a), irradiation dose produces a progressive reduction in fatigue life, with the magnitude of degradation becoming more pronounced at higher strain amplitudes, reflected by a gradual steepening of the fatigue-life surface. In contrast, EUROFER97 (Fig. 7d) shows comparatively weak sensitivity to irradiation dose, with nearly parallel fatigue-life contours along the dose axis across the entire strain range, indicating a largely uniform irradiation response.

The coupled effects of irradiation dose and test temperature at a fixed strain amplitude of 1% are shown in Figs. 8b and 8e. For SS316 (Fig. 7b), fatigue life decreases smoothly with increasing dose and temperature, forming a gently sloped surface that suggests a predominantly additive influence of these parameters. EUROFER97 (Fig. 7e), however, exhibits a markedly different response: while fatigue life remains relatively insensitive to irradiation dose, a sharp drop in the number of cycles is observed as the test temperature exceeds ~550 °C. This is manifested as pronounced curvature in the fatigue-life surface along the temperature axis, consistent with the onset of microstructural instability when the testing temperature approaches and exceeds the alloy's tempering heat-treatment temperature, leading to degradation of the tempered martensitic structure and reduced fatigue resistance.

Figs. 7c and 7f depict the interaction between strain amplitude and test temperature at a constant irradiation dose of 30 dpa. In SS316 (Fig. 7c), increasing temperature leads to a systematic reduction in fatigue life across all strain amplitudes, with temperature effects becoming increasingly significant at higher strain levels. In EUROFER97 (Fig. 7f), fatigue life remains strongly governed by strain amplitude at lower and intermediate temperatures; however, at higher strain amplitudes and higher temperatures, particularly beyond ~550 °C, a pronounced decline in fatigue life is again evident, consistent with the temperature-dominated degradation observed in Fig. 7e.

Overall, the multivariate analysis reveals distinct degradation paradigms. SS316 exhibits continuous sensitivity to strain amplitudes, irradiation dose, and temperature, with combined effects leading to progressively reduced fatigue life. EUROFER97 exhibits limited sensitivity to irradiation dose but a pronounced temperature dependence, which becomes significant beyond a critical threshold, above which fatigue life deteriorates rapidly due to microstructural instability. These results highlight the importance of temperature limits on the fatigue performance of EUROFER97 under irradiation and underscore the PINN model's ability to capture these physically meaningful interaction effects under reactor-relevant conditions.

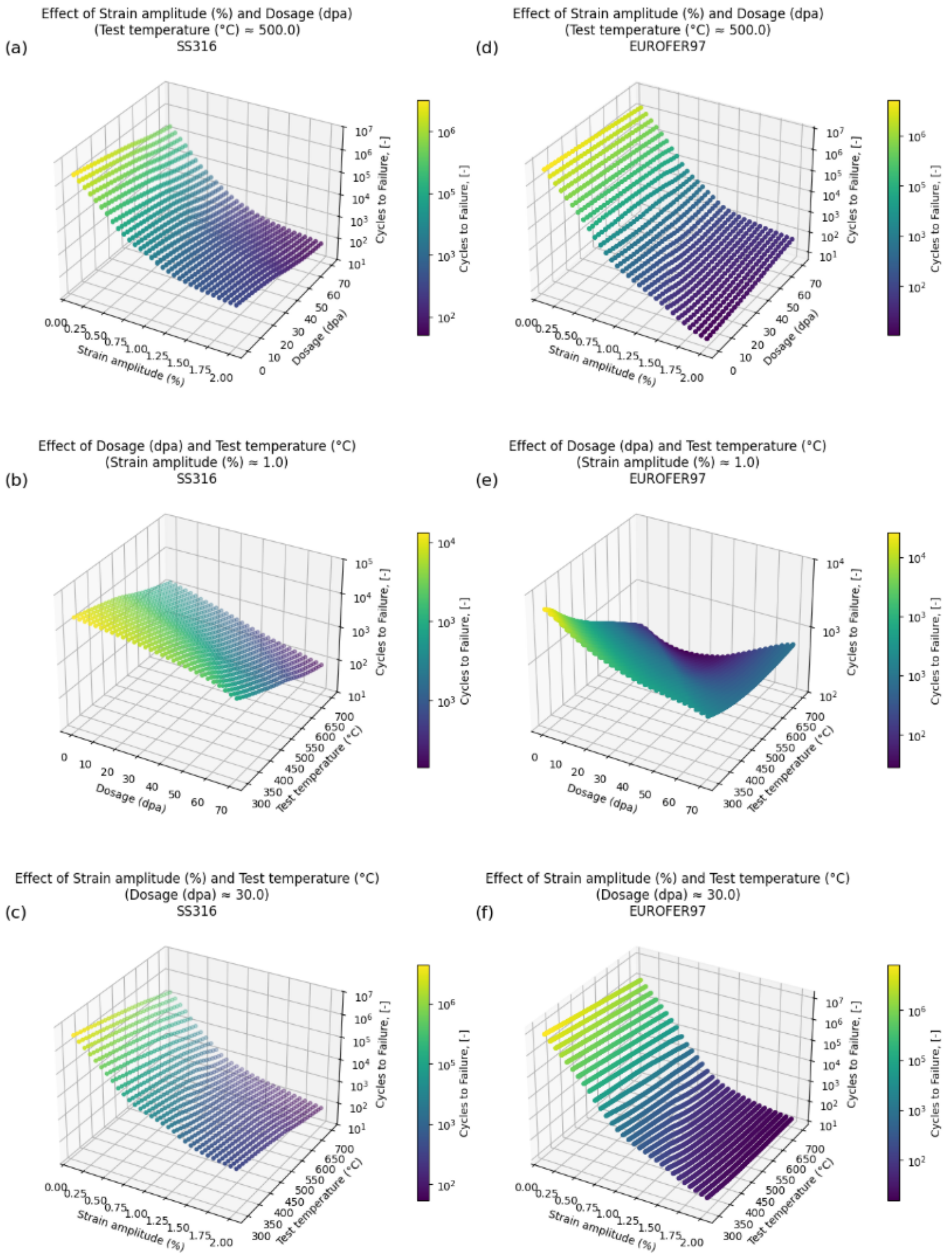

Fig. 7: Predicted three-dimensional fatigue life surfaces illustrating the coupled influence of strain amplitude, irradiation dose, and test temperature on fatigue behaviour of (a–c) austenitic SS316 and (d–f) ferritic/martensitic (F/M) EUROFER97 steels, as obtained from the PINN model. In (a) and (d), the test temperature is fixed at 500 °C while the strain amplitude and irradiation dose are varied. In (b) and (e), the strain amplitude is held constant at 1.0 % and the combined effect of irradiation dose (0–70 dpa) and test temperature (300–700 °C) is examined. In (c) and (f), the irradiation dose is fixed at 30 dpa, and the influence of strain amplitude and test temperature is shown. For both alloys, fatigue life decreases monotonically with increasing strain amplitude; however, SS316 exhibits a more pronounced and progressive degradation with increasing dose and temperature, whereas EUROFER97 shows a comparatively milder response, highlighting its superior resistance to fatigue damage.

### 3.6. Limitations and Outlook

The effectiveness of any ML framework depends on the size, quality, and representativeness of the underlying dataset. In the present study, the available dataset on irradiation fatigue is limited. Moreover, irradiation parameters inherently involve assumptions and measurement uncertainties, resulting in data scatter. Additional variability arises from differences in irradiation spectrum, flux, specimen geometry, and post-irradiation testing protocols, introducing heterogeneity that cannot be fully separated within the current dataset. Moreover, critical microstructural descriptors, such as defect density, precipitate evolution, and dislocation structures governing irradiation hardening and fatigue behavior, are not included, primarily because such detailed information is rarely reported in the open literature. As a result, the learned relationships remain predominantly phenomenological rather than fully mechanistic.

It is important to emphasize that the univariate and multivariate analyses shown in Figs. 6 and 7 are performed within the global parameter envelope of the training dataset. However, the dataset is highly non-uniform across alloys, with each alloy occupying only a limited subregion of the feature space. Consequently, several predictions correspond to conditions beyond the experimentally explored range for specific alloys, representing cross-alloy extrapolation, i.e., predicting behaviour of a given alloy in regimes where that alloy has no direct experimental data, while leveraging information learned from other alloys. While the physics-informed framework promotes physically consistent and reasonable trends, these predictions do not substitute for independent experimental validation, particularly at high irradiation doses and elevated temperatures, where data remain sparse.

Although physics-based constraints improve generalization compared to purely data-driven models, reliability outside the training domain cannot be guaranteed. Therefore, predictions in such extreme regimes should be interpreted with caution until further high-dose and high-temperature experimental data become available. Moreover, the imposed physics constraints are simplified representations of the complex interactions governing irradiation-assisted fatigue. For example, monotonic degradation priors may not fully capture regime-dependent effects such as irradiation hardening or strain-amplitude-sensitive life extension. While implemented as soft constraints, their formulation can be further refined, and additional mechanistic fatigue laws may be incorporated to better represent condition-dependent behavior.

Future research incorporating larger and more comprehensive datasets, including both mechanical and microstructural variables, along with refined physics-based constraints that account for irradiation hardening and competing damage mechanisms, holds strong potential to further enhance the robustness, mechanistic fidelity, and predictive reliability of the proposed framework.

## 4. Summary and conclusions

This study presents the first physics-informed neural network (PINN) framework for predicting LCF life in neutron-irradiated austenitic and F/M steels, significantly reducing reliance on extensive post-irradiation testing. The model was trained on a curated dataset of 495 strain-controlled fatigue data points, spanning both irradiated and unirradiated conditions, and incorporating 50 literature-derived input features. By embedding physical constraints through partial differential equations within the loss function, the PINN ensures physically consistent learning while improving predictive accuracy. Compared with conventional machine-learning models (RF, GB, XGB, and NN), the PINN consistently achieved superior performance, with the highest average $R^2$ value (~0.85) and the lowest average MSE value (~0.07) across multiple data splits. SHAP-based interpretability analysis identified strain amplitude, irradiation dose, and test temperature as the most influential parameters, each exhibiting physically meaningful inverse correlations with fatigue life.

Univariate analyses confirmed strain amplitude as the dominant driver of fatigue degradation for both steel classes, while irradiation dose and temperature introduced material-dependent sensitivities. The model reproduced established experimental trends, including the greater irradiation tolerance and dose saturation behavior of F/M steels relative to austenitic SS316, consistent with enhanced defect recombination and reduced void swelling in F/M microstructures. Multivariate analyses revealed distinct interaction effects. SS316 exhibited strong coupling among strain amplitude, irradiation dose, and temperature, resulting in nonlinear fatigue degradation under combined loading. In contrast, EUROFER97 showed weak sensitivity to irradiation dose but a pronounced reduction in fatigue life above ~550 °C, associated with microstructural instability when testing temperatures exceed its tempering heat-treatment temperature. These results highlight fundamentally different degradation paradigms in austenitic and F/M steels and demonstrate the PINN model's capability to capture physically meaningful interaction effects.

Despite these advances, the framework's reliability is influenced by a limited, non-uniform dataset and the absence of quantitative microstructural descriptors, which are rarely available in literature. While physics-informed constraints enhance generalization, predictions at extreme irradiation doses and temperatures require cautious interpretation. Future efforts incorporating larger datasets enriched with microstructural descriptors, along with refined mechanistic constraints, will be essential to enhance predictive robustness and mechanistic fidelity.

**Acknowledgements**

AC gratefully acknowledges the financial support from the Indian Institute of Science (IISc) and the Infosys Foundation, Bengaluru.

## Supplementary Materials

### Physics-informed neural network for predicting fatigue life of unirradiated and irradiated austenitic and ferritic/martensitic steels under reactor-relevant conditions

Dhiraj S. Kori[1], Abhinav Chandraker[1], Syed Abdur Rahman[2], Punit Rathore[3] and Ankur Chauhan[1]

[1]Extreme Environments Materials Group (EEMG), Department of Materials Engineering, Indian Institute of Science (IISc), Bengaluru, 560012 Karnataka, India

[2]Department of Materials, University of Oxford, Parks Road, Oxford, OX1 3PH, United Kingdom

[3]Robert Bosch Centre for Cyberphysical Systems, Indian Institute of Science (IISc), Bengaluru, 560012 Karnataka, India

*Corresponding author: ankurchauhan@iisc.ac.in

### S.1: Data Composition of all the steels considered in this work from the literature.

Table S.1: Composition of all the steels considered in this work from the literature [1–12]. This includes 22 input features incorporating compositional variables.

| Alloy | Element (Range (%)) |
|---|---|
| SS304 | C (0–0.08), Mn (0–2), P (0–0.045), S (0–0.03), Si (0–1), Cr (18–20), Ni (8–10.5), N (0–0.1) |
| SS304L | C (0–0.03), Mn (0–2), P (0–0.045), S (0–0.03), Si (0–1), Cr (18–20), Ni (8–12), N (0–0.1) |
| SS316 | C (0.03–0.08), Mn (0–2), P (0–0.045), S (0–0.03), Si (0–1), Cr (16–18), Ni (10–14), Mo (2–3), N (0–0.1), Co (0.03) |
| SS316L | C (0–0.03), Mn (0–2), P (0–0.045), S (0–0.03), Si (0–1), Cr (16–18), Ni (10–14), Mo (2–3), N (0–0.1), Co (0.17) |
| SS310 | C (0–0.25), Mn (0–2), P (0–0.045), S (0–0.03), Si (0–1.5), Cr (24–26), Ni (19–22), Co (0.076) |
| F82H | C (0.08–0.12), Mn (0.05–0.21), P (0–0.01), S (0–0.01), Si (0.05–0.2), Cr (7.4–8.5), Ni (0–0.1), V (0.15–0.25), N (0–0.02), W (1.8–2.2), Ta (0.01–0.06), Mo (0–0.05), Ti (0–0.012), Co (0-0.005) |
| EUROFER97 | C (0.09–0.12), Mn (0.22–0.61), P (0–0.01), S (0–0.005), Si (0–0.1), Cr (8.5–9.5), Ni (0–0.05), V (0.15–0.25), W (1–1.3), Ta (0.055–0.95), Mo (0–0.05), Ti (0–0.01), Y2O3 (0-0.5), Co (0-0.006) |
| T91 | C (0.07–0.14), Mn (0.3–0.6), P (0–0.02), S (0–0.01), Cr (8–9.5), Ni (0–0.4), Mo (0.8–1.05), V (0.18–0.25), Si (0–0.25), Al (0–0.013), Ti (0–0.01), Nb (0.06–0.1), Co (0.013) |
| HT-9 | C (0.17–0.23), Mn (0.4–0.7), Si (0.1–0.3), Cr (11–12.5), Mo (0.8–1.2), V (0.25–0.35), W (0.4–0.6), Co (0.015-0.021) |
| OPTIFER IVc | C (0.1–0.13), Mn (0.42–0.6), P (0–0.01), S (0–0.005), Si (0–0.155), Cr (9.25–10.75), Ni (0–0.009), V (0.18–0.3), W (1–1.2), Ta (0.05–0.95), Mo (0–0.005), Ti (0–0.01) |

Table S.2: Range of values for the input features [1–12]. There are 16 input features with defined ranges and 12 with category encoding.

| Feature | Min Value | Max Value |
|---|---|---|
| Normalizing Temperature (°C) | 950 | 1050 |
| Normalizing Time (min) | 10 | 60 |
| Tempering Temperature (°C) | 700 | 780 |
| Tempering Time (min) | 30 | 300 |
| Hot Rolling Temperature (°C) | 1050 | 1200 |
| Annealing Temperature (°C) | 929 | 1070 |
| Annealing Time (min) | 15 | 60 |
| Additional Pre-treatments (Category encoded) | Solution-annealed | |
| | HS-CS | |
| | Cold worked (CW)- 13 % | |
| | CST | |
| | Cold worked (CW) - 11 % | |
| | Austenitized | |
| Irradiation Environment (Category encoded) | He-Ne | |
| | Ar | |
| | Static Na | |
| | Flowing Na | |
| Dosage (dpa) | 0 | 71 |
| Test temperature (°C) | 22 | 700 |
| Strain amplitude (%) | 0.14 | 6.15 |
| Cycles to failure | 11 | 247907 |
| Sample Type<br><br>(Category encoded) | Hourglass<br><br>Cylinder | |
| Minimum Diameter (mm) | 1.25 | 5.1 |
| Gauge Length (mm) | 6 | 25.6 |
| Radius of Curvature (mm) | 10 | inf |
| Strain rate ($s^{-1}$) | 0.003 | 0.8 |
| Irradiation Temperature (°C) | 55 | 750 |

**S.2: Correlation analysis between Experimental Input Features and Fatigue life**

To examine the direct relationships between individual experimental parameters and fatigue life, log–log scatter plots of (a) strain amplitude, (b) test temperature, and (c) irradiation dose versus fatigue life are presented. A clear monotonic trend with strain amplitude is observed, consistent with established fatigue behavior. In contrast, no direct correlation is evident for temperature and irradiation dose. This apparent absence of a trend arises from multivariate coupling, where the simultaneous variation of parameters obscures individual effects in a two-dimensional projection. For instance, a low-strain test conducted at elevated temperature may exhibit a longer fatigue life than a high-strain test performed at lower temperature, resulting in scatter that masks the underlying physical relationships. Consequently, simple scatter plots are insufficient to fully capture the interdependencies among the governing variables.

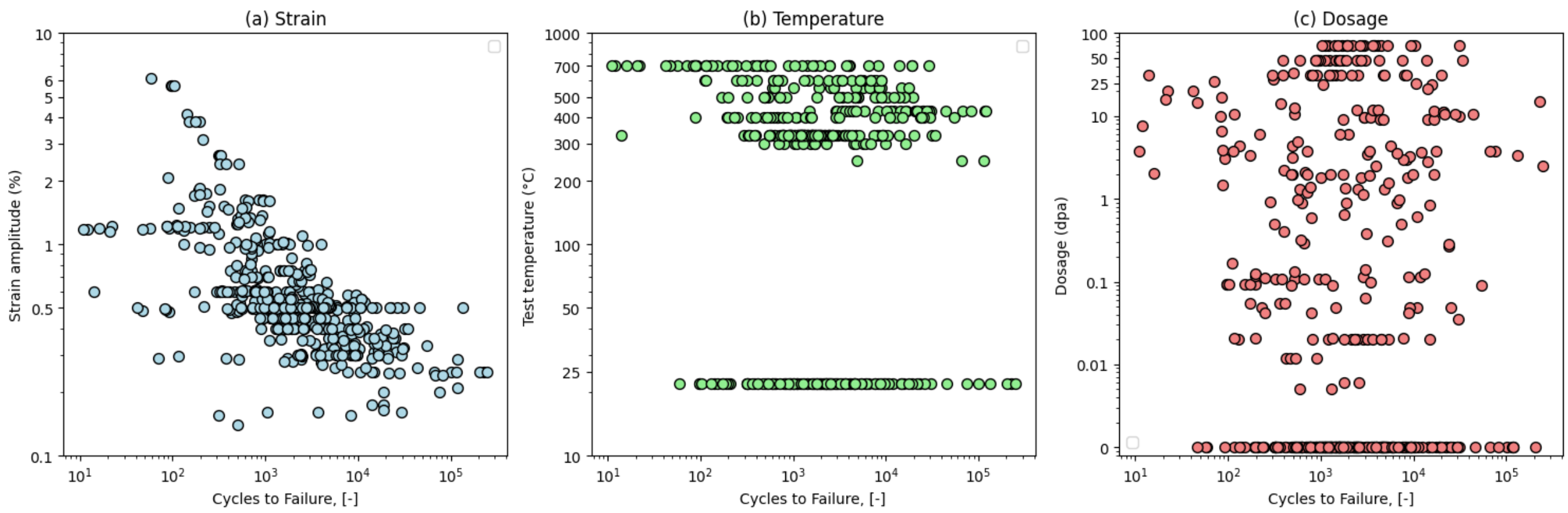

Fig. S.1: Plots showing log-log scatter plots for (a) Strain amplitude (%), (b) Test temperature (°C), and (c) Dosage (dpa) vs Number of cycles to failure, along with the general linear regression trendline representing a power-law fit to show their monotonic relationship. The zero-dose values were set to 10-3 to represent them in the plot.

Additionally, a boxplot of the normalized input features is presented to illustrate their statistical distributions. The central line indicates the median, the box represents the interquartile range (25th–75th percentiles), whiskers denote the typical data range, and individual points correspond to statistical outliers. The figure below shows the normalized distributions of strain amplitude, test temperature, and irradiation dose. Strain amplitude and dose exhibit pronounced right-skewness with heavy tails, whereas temperature is more uniformly distributed. The irradiation dose distribution is dominated by low-dose data with only a few high-dose samples, reflecting the experimental challenges associated with irradiated fatigue testing. This distribution motivates the use of monotonic physics-based regularization to stabilize model learning in data-scarce regimes, rather than relying solely on data-driven extrapolation.

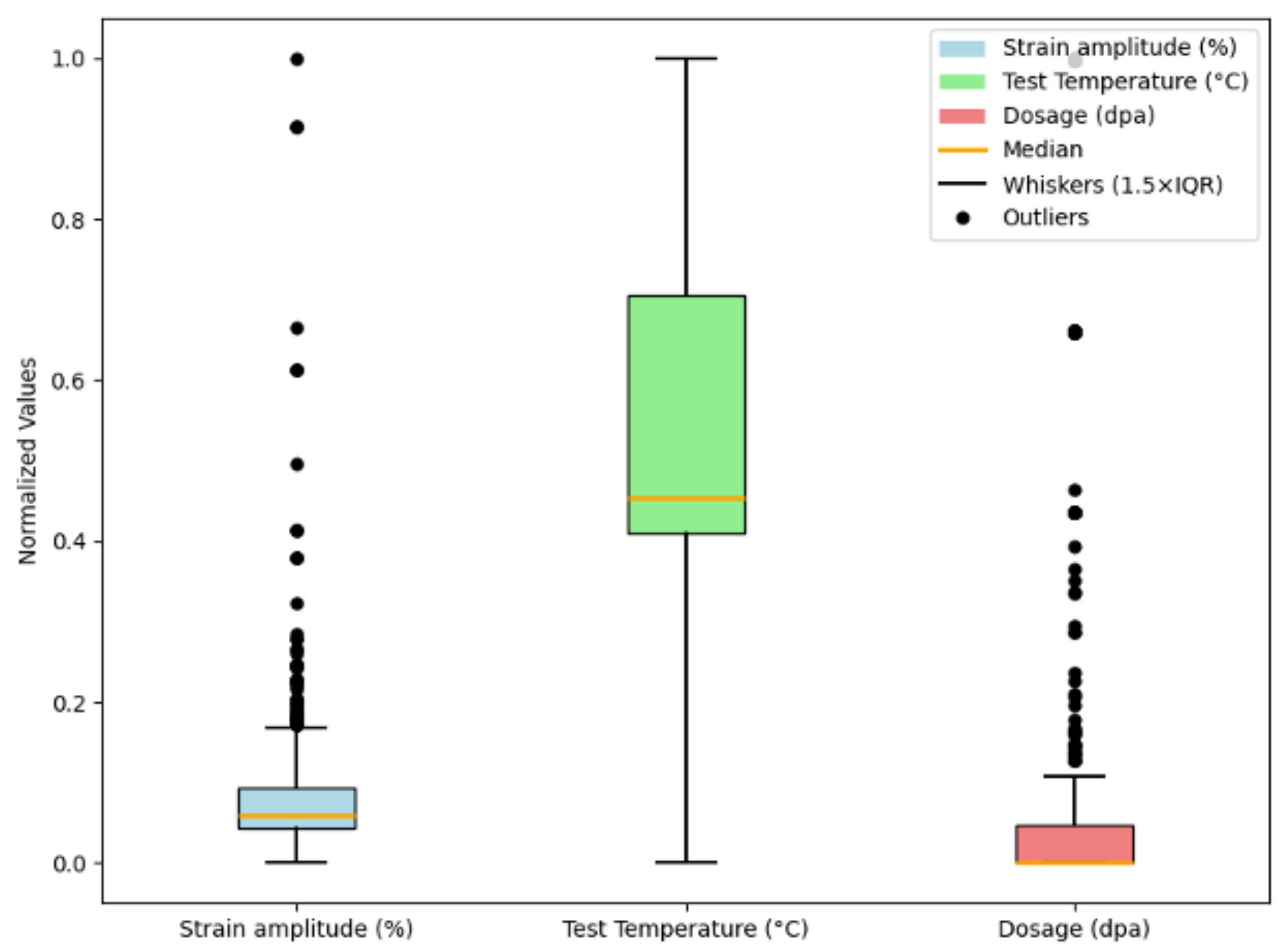

Fig. S.2: Box plot showing the normalized distributions (scaled from 0 to 1) of strain amplitude (%), test temperature (°C), and irradiation dose (dpa). The orange line indicates the median, the boxes represent the interquartile range (IQR, 25%–75%), and the whiskers denote the spread of the data excluding outliers, which are shown as black circles. The distributions of strain amplitude and irradiation dose are strongly skewed, with limited coverage at higher values, whereas the test temperature distribution is comparatively broader. This non-uniform distribution highlights the challenges of purely data-driven learning in sparsely sampled regions and motivates the use of physics-informed monotonic regularization to stabilize predictions beyond densely populated areas of the feature space.

### S.3: Handling Infinity and Missing values

During pre-processing, missing and infinite normalized values were replaced with extreme placeholder values (±10), which may appear unconventional. To ensure that this choice does not significantly influence model performance, a sensitivity analysis was performed by varying the replacement values to ±2, ±5, ±10, and ±100. As shown in the figure below, the predictive performance remains stable for moderate replacement values (±2, ±5, ±10), with $R^2$ ranging from 0.82 to 0.85 and MSE between 0.08 and 0.09. In contrast, using an extremely large value such as ±100 results in a noticeable reduction in accuracy ($R^2 = 0.698$ and MSE = 0.157) and leads to unstable convergence. These results indicate that using ±10 during preprocessing does not affect the model conclusions, whereas excessively large values such as ±100 can degrade performance.

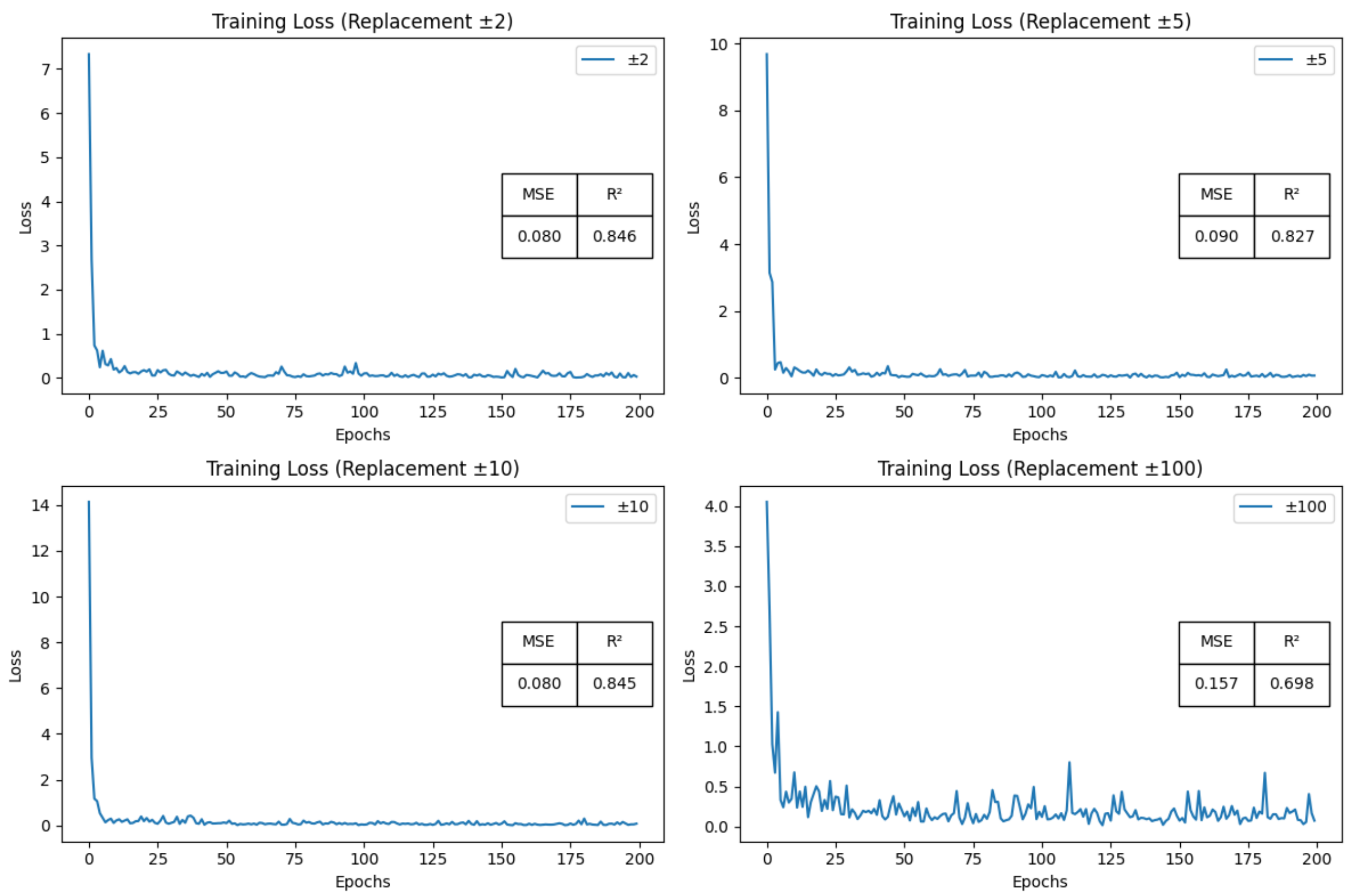

Fig. S.3: Plots showing sensitivity analysis done by varying the replacement values for infinity and missing values in the normalized dataset. While ±2, ±5, and ±10 values show relatively stable predictive performance, ±100 shows a noticeable decrease in accuracy and unstable convergence.

## S.4: Brief descriptions of the employed ML algorithms

### S.4.1: Artificial Neural Networks

Artificial Neural Networks (ANNs) are a class of machine learning algorithms inspired by the structure and function of the human brain [13]. They consist of interconnected layers of artificial neurons that process input data and produce output predictions. ANNs comprise input, hidden, and output layers (see Fig. S.1). The input layer receives the initial feature vector and transmits it to the hidden layers, where a weighted sum of the input data is performed. The output layer generates predictions or classifications. Training involves iterative processes that adjust the weights and biases, enabling accurate predictions on unseen data.

A feedforward NN architecture was defined using the library PyTorch [14]. Some of its key components are described:

- **Input Layer:** The network takes the input of dimension input_dim (the number of neurons)
- **Hidden Layers:** It consists of two hidden layers, each with dimensions hidden_dim1 and hidden_dim2 (the number of neurons). Increasing the values of the dimensions allows the network to learn more complex patterns in the data, but it also increases the risk of overfitting if not properly regularised.

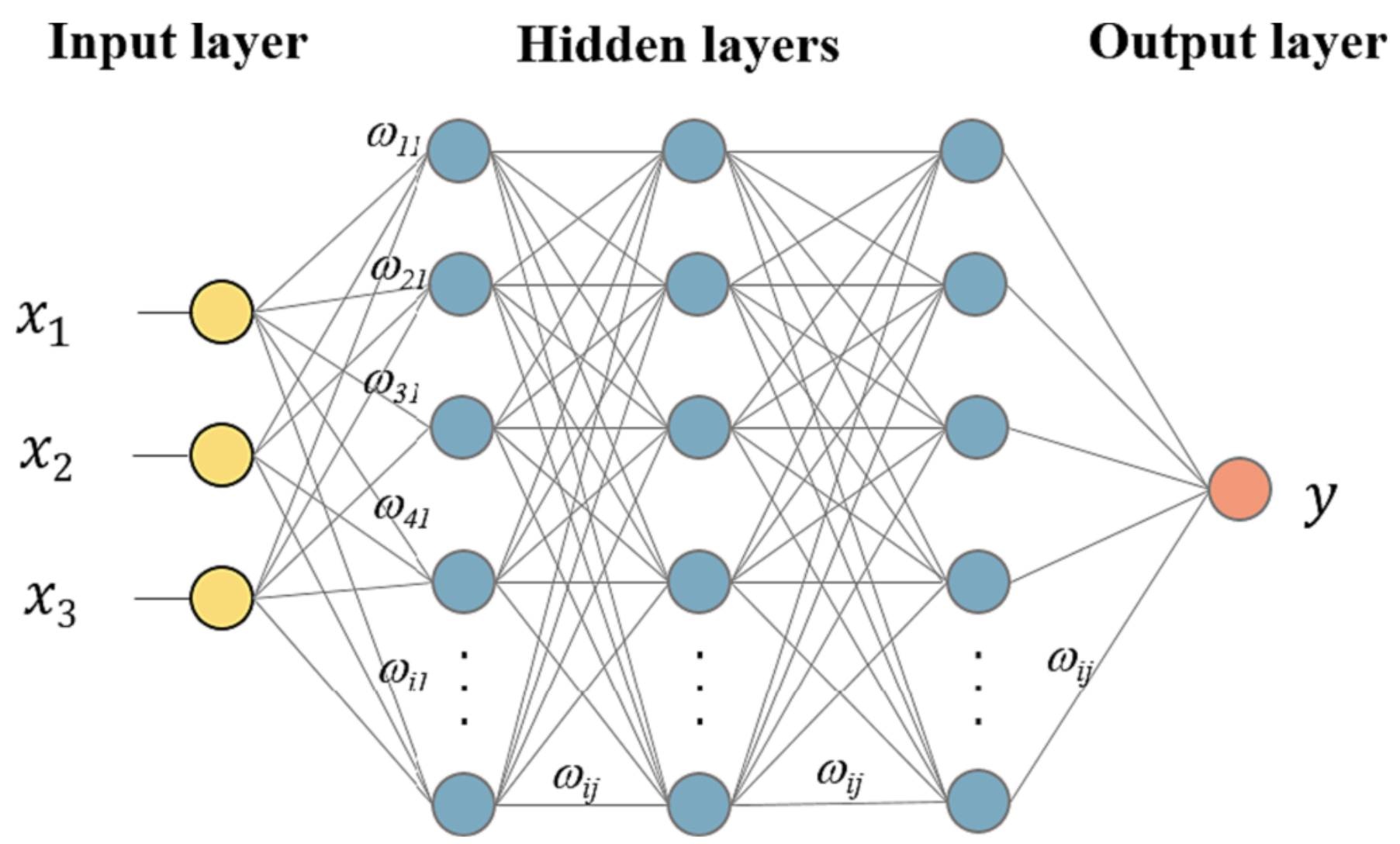

Fig. S.4: Architecture of a Neural Network showcasing Input, Hidden, and Output Layers.

- **Activation Function:** The activation function used between layers introduces nonlinearity, thereby preserving the data's mean and variance during training, potentially reducing the need for batch normalization. Some of the activation functions used are the Hyperbolic Tangent (Tanh), sigmoid linear unit (SiLU), exponential linear unit (ELU), rectified linear unit (ReLU), and Gaussian error linear unit (GELU). The activation function used in this model is the Scaled Exponential Linear Unit (SELU) (see Fig. S2), which helps in faster convergence [15].

$$SELU(x) = \lambda \begin{cases} x, & x > 0 \\ \alpha(e^{x} - 1), & x \leq 0 \end{cases}$$

Where:

$\lambda \approx 1.05074$

$\alpha \approx 1.67326$

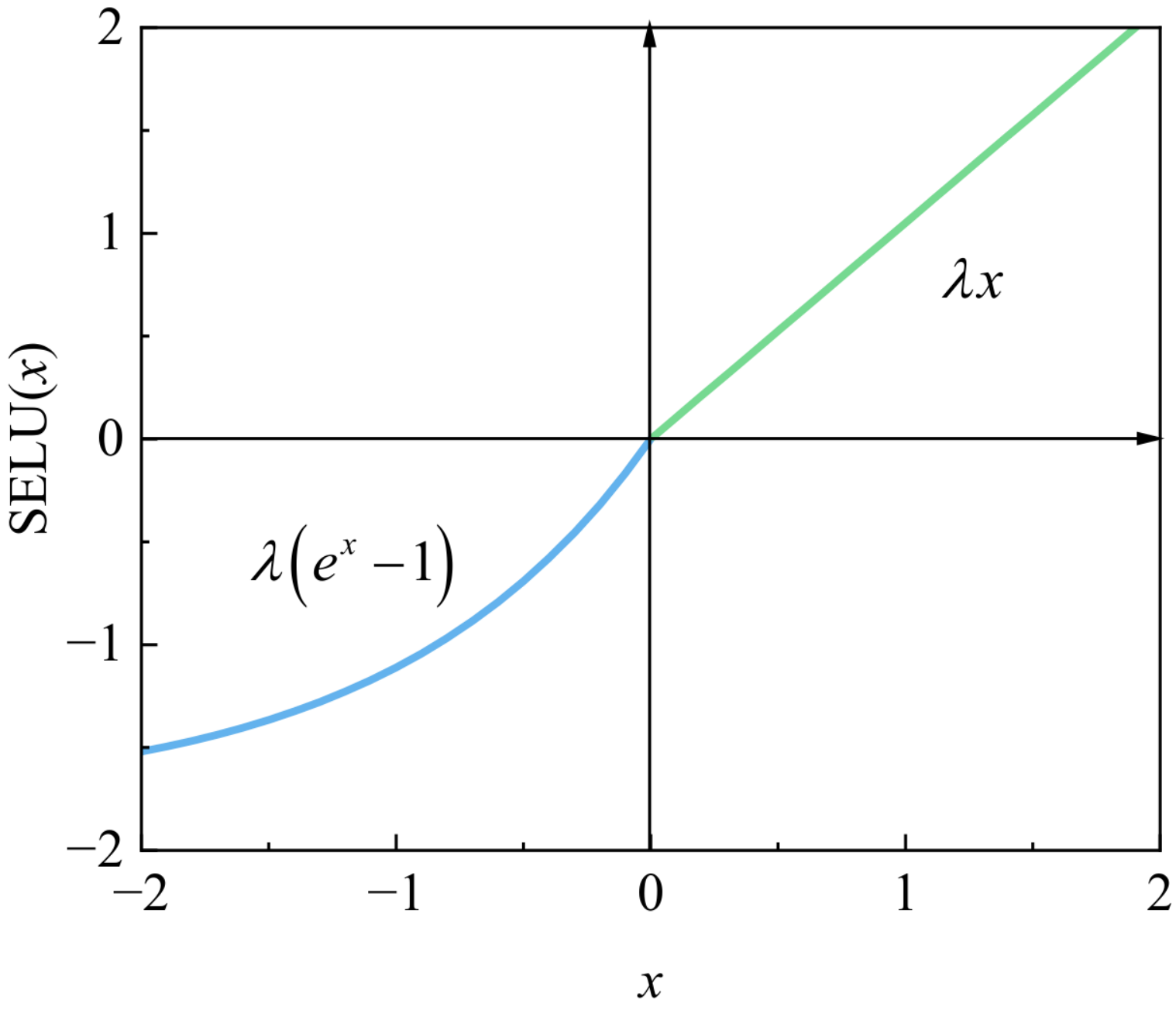

Fig. S.5: Graph of the activation function, SELU(x) vs x [16]. X is the output of the hidden layer for the activation function, SELU (x).

- **Output Layer:** The output layer consists of a single neuron that represents the predicted value. The calculation process in each neuron can be represented as:

$$a = f\left(\sum (w \times x) + b\right)$$

Where:

a: Output

w: Weight

x: Input

b: Bias

f: Activation Function

For the entire NN, the process of summing the weighted inputs of neurons can be represented as:

$$z_j^{(n)} = \sum_i \left(w_{j\times i}^{(n)} \cdot x_i\right) + b_j^{(n)}$$

$$Z_{j\times1}^{(n)}=\begin{cases} W_{j\times1}^{(1)}\times X_{i\times1}+B_{j\times1}^{(1)}, & n=1 \\ W_{j\times1}^{(n)}\times A_{j\times1}^{(n\text{-}1)}+B_{j\times1}^{(n)}, & n>1 \end{cases}$$

Where:

i: Number of neurons in the previous layer. If the hidden layer is considered first, it represents the number of input features.

j: Number of neurons in current layer

n: Level of the hidden layer

$X_{i\times1}=\begin{bmatrix} x_1 \\ x_2 \\ \vdots \\ x_i \end{bmatrix}$: Input layer matrix

$W_{j\times i}^{(n)}=\begin{bmatrix} w_{1\times1}^{(n)} & w_{1\times2}^{(n)} & \cdots & w_{1\times i}^{(n)} \\ w_{2\times1}^{(n)} & w_{2\times2}^{(n)} & \cdots & w_{2\times i}^{(n)} \\ \vdots & \vdots & \ddots & \vdots \\ w_{j\times1}^{(n)} & w_{j\times2}^{(n)} & \cdots & w_{j\times i}^{(n)} \end{bmatrix}$: Weight matrix of $n^{th}$ hidden layer

$B_{j\times1}^{(n)}=\begin{bmatrix} b_1^{(n)} \\ b_2^{(n)} \\ \vdots \\ b_j^{(n)} \end{bmatrix}$: Bias matrix of $n^{th}$ hidden layer

The output values of the $n^{th}$ hidden layer after passing through an activation function are:

$$a_j^{(n\text{-}1)}=f_{n\text{-}1}\left(z_j^{(n\text{-}1)}\right)$$

$$A_{j\times1}^{(n\text{-}1)}=f_{n\text{-}1}\left(Z_{j\times1}^{(n\text{-}1)}\right)$$

Where $A_{j\times1}^{(n)}$ is the weighted Output of the $n^{th}$ hidden layer, and $Z_{j\times1}^{(n)}$ is the weighted sum of the $n^{th}$ hidden layer

Final Output of the NN, Y is given by:

$$y=f_m\left(\sum_i\left(w_{m\times i}^{(k)}\cdot a_i\right)+b_m^{(k)}\right)$$

$$Y = f_m\left(Z_{k\times 1}^{(m)}\right)$$

Where m is the number of layers in the final hidden layer and k is the number of outputs

### S.4.2: Random Forests:

Random Forest (RF) is a robust ensemble learning algorithm that builds multiple decision trees (DTs) to make predictions. It follows the *divide-and-conquer* strategy to enhance predictive performance and address the overfitting issues commonly associated with using a single decision tree [17]. DTs partition the feature space into regions based on simple decision rules, and each DT in a Random Forest comprises *decision nodes*, which apply decision-making rules, and *leaf nodes*, which contain the final outcomes [18]. RF uses a technique called *bootstrapping to ensure diversity among the DTs*, where each tree is trained on a different subset of the training data sampled *with replacement*. This process introduces randomness and reduces model variance.

For regression tasks, the final output is obtained through a voting mechanism or by averaging the predictions from all individual trees. This averaging mechanism helps reduce overfitting by smoothing predictions and mitigating the influence of a single overfit tree.

Key hyperparameters in Random Forest that influence its complexity and performance include:

- **Maximum depth** of the trees (controls how deep each tree can grow),
- **Maximum number of features** considered for splitting at each decision node (introduces randomness and prevents dominant features from being overused),
- **Number of trees** in the ensemble (controls the model's capacity and stability).

By tuning these hyperparameters appropriately, the RF algorithm can achieve high accuracy while maintaining generalization across unseen data.

### S.4.3: Gradient Boosting:

Gradient Boosting (GB) is an ensemble-based algorithm that sequentially constructs models using DTs and differs fundamentally from RF. While RF builds multiple deep and independent trees in parallel, GB constructs relatively shallow DTs, also known as 'Weak Learners', in sequence, creating a single 'Strong Learner', with each tree attempting to correct the errors of its predecessor [18].

The GB process begins by computing the average of the target values across the training data. This average serves as the initial prediction for all data points. The algorithm then calculates the differences between actual and predicted values, called as the loss function, and trains the first decision tree to fit them. The predicted output is then updated by adding the contribution of this tree (scaled by a *learning rate*) to the initial prediction.

This procedure is repeated iteratively: in each step, a new decision tree is trained to predict the residual errors from the cumulative prediction of the previous trees. In doing so, the model gradually reduces the loss function, ultimately improving predictive performance.

Key hyperparameters that influence the behaviour and performance of GB include:

- **Maximum depth** of each tree (controls the complexity of individual learners),
- **Learning rate** (scales each tree's contribution to the final prediction),
- **Number of trees** (determines the number of iterations the boosting process runs).

The final prediction for each data point is obtained by aggregating the initial average with the weighted sum of all subsequent tree outputs. Mathematically, the final model output can be expressed as:

$$FinalPrediction = \overline{y} + \sum_{i=1}^{N} \eta \cdot h_i(x)$$

Where $\overline{y}$ is the average of the actual outputs (initial prediction), $\eta$ is the learning rate, $h_i(x)$ is the prediction from the $i$-*th* decision tree, and $N$ is the total number of trees.

**S.4.4: eXtreme Gradient Boosting:**

**eXtreme Gradient Boosting (XGB)** is an advanced ensemble algorithm that builds upon the foundational principles of GB. While both algorithms operate by sequentially correcting errors from previous iterations, XGB introduces several enhancements in regularization, efficiency, and model structure.

Unlike traditional DTs, the trees in XGB employ a more sophisticated mechanism for node evaluation by incorporating the concept of the *Amount of Say*. This metric quantifies each node's influence or importance in the decision-making process and plays a central role in determining optimal splits.

The *Amount of Say* at each decision node is computed using the formula:

$$AmountofSay = \frac{(\sum g_i)^2}{\sum h_i + \lambda}$$

Where $g_i$ is the first-order gradient (loss derivative), $h_i$ is the second-order gradient (Hessian), and $\lambda$ is the regularization parameter to control complexity.

The **gain** at a decision node, which measures the improvement in the loss function from a potential split, is given by:

$$Gain = AmountofSay \quad + AmountofSay_{child} - AmountofSay_{parentnode}$$

The algorithm selects splits that yield the highest gain, effectively focusing on nodes that lead to the greatest reduction in loss.

XGB introduces several important **hyperparameters** to control the model's learning and complexity:

- $\eta$: learning rate (controls step size in updates),
- $\lambda$: $L^2$ regularization term (prevents overfitting),
- $\gamma$: minimum loss reduction required to make a further partition on a leaf node (pruning parameter),
- max_depth: maximum depth of the trees,
- n_estimators: number of boosting rounds (trees in the ensemble).

Together, these improvements make XGB faster, more accurate, and more robust than standard GB.

### S.4.5: SHapley Additive exPlanations

SHapley Additive exPlanations (SHAP) is a model-agnostic interpretability technique derived from cooperative game theory that quantifies the contribution of each input feature to a machine learning model's prediction [19]. It provides a principled way to understand the importance and influence of individual input features on the model's output, both globally (across the entire data) and locally (for each instance).

For any given prediction, SHAP assigns a SHAP value to every input feature. A larger SHAP value indicates a stronger influence of that feature on the output prediction, either positively or negatively. Importantly, SHAP does not just assess feature importance globally; it also enables instance-level interpretability, allowing us to understand *how* each input feature contributed to a specific prediction.

To compute the SHAP value of a target feature for a given instance, the algorithm considers all possible subsets of input features that include the target feature. The model is evaluated on each of these subsets to estimate its predictive accuracy. Then, the target feature is removed, and the model is re-evaluated on the reduced subset. The difference in prediction reflects the marginal contribution of the target feature. SHAP aggregates these differences across all subsets using a weighted average based on each subset's size.

Mathematically, the SHAP value for feature $i$ with model $f$ and instance $x$ is defined as:

$$\Phi_i(f,x)=\sum_{S\subseteq[F\text{-}\{i\}]}\frac{|S|!\,(M\text{-}|S|\text{-}1)!}{M!}[f_x(S\cup\{i\})\text{-}f_x(S)]$$

Where $\Phi_{i(f,x)}$ is the SHAP value for feature $i$, S is a subset of the input feature set F, excluding feature i, M is the total number of input features, $f(S)$ is the model prediction using only the features in subset S, and $f(S\cup\{i\})$ is the prediction when feature i is added to subset S.

This equation ensures that each feature’s contribution is distributed fairly based on its marginal impact across *all possible combinations*, aligning with the principles of Shapley values from cooperative game theory.

**S.5: Derivation of First- and Second-Order Derivatives in the PINN**

In the implementation of the PINN, evaluation of the PDE residual requires computation of both the first and second-order derivatives of the network output y with respect to the input variables $x_i$. Since y is obtained through a composition of multiple neural network layers, these derivatives are computed using repeated application of the chain rule across all layers.

Let $z^{(l)}$ denote the pre-activation vector at the l-th layer, and $a^{(l)}$ is activation output. For clarity, neuron indices at each layer are denoted as m, n, p, and q. The forward pass of the network can be represented as:

$$x \longrightarrow z^{(1)} \longrightarrow a^{(1)} \longrightarrow z^{(2)} \longrightarrow a^{(2)} \cdots \ \longrightarrow z^{(l)} \longrightarrow a^{(l)} \longrightarrow \cdots \longrightarrow z^{(k)} \longrightarrow y$$

Where

$$z_m^{(l)} = \sum_i \left( w_{m \times i}^{(l)} \cdot a_i^{(l\text{-}1)} \right) + b_m^{(l)}$$

$$a^{(l)} = f^{(l)}\left(z^{(l)}\right) ; \ a^{(0)} = x \, ; a^{(k)} = y$$

Here, each summation accounts for the fact that all neurons in each layer contribute to the derivative via their weighted connections, which means that y is a nested composition. The first partial derivative of y with respect to an input $x_i$ is obtained by propagating sensitivities backward through the network. For derivative computations, we use Jacobian matrices.

Let:

$$J_f(x) = \frac{\partial f}{\partial x} = \left[ \frac{\partial f}{\partial x_1}, \frac{\partial f}{\partial x_2}, \cdots\cdots, \frac{\partial f}{\partial x_n} \right]$$

Starting from the output layer, we get:

$$\frac{\partial y}{\partial x_i} = \nabla_{x_i} y = \frac{\partial a^{(k)}}{\partial z^{(k)}} \frac{\partial z^{(k)}}{\partial x_i} = J_{a^{(k)}}\left(z^{(k)}\right) J_{z^{(k)}}(x_i)$$

The output layer's pre-activation $z_m^{(k)}$ depends on the activations from the previous layers $a_n^{(k\text{-}1)}$:

$$J_{z^{(k)}}(x_i) = \frac{\partial z_m^{(k)}}{\partial x_i} = \sum_n \frac{\partial z_m^{(k)}}{\partial a_n^{(k\text{-}1)}} \frac{\partial a_n^{(k\text{-}1)}}{\partial x_i} = J_{z^{(k)}}\left(a^{(k-1)}\right) J_{a^{(k-1)}}(x_i)$$

But;

$$J_{a^{(k)}}(x_i) \ = \frac{\partial a_n^{(k)}}{\partial x_i} = \frac{\partial a_n^{(k)}}{\partial z_n^{(k)}} \frac{\partial z_n^{(k)}}{\partial x_i} = J_{a^{(k)}}\left(z^{(k)}\right) J_{z^{(k)}}(x_i)$$

Repeating this process until we reach the input layer, we get the final expression for the first derivative:

$$\nabla_{x_i} y = \prod_{l=k}^{1} \left[ J_{a^{(l)}}\left(z^{(l)}\right) J_{z^{(l)}}\left(a^{(l-1)}\right) \right]$$

Further differentiating the first derivative to get the second derivative:

$$\nabla_{x_i}^2 y = \sum_{p=k}^{1} \left[ F_{k,p} \left[ H_p(x_i) \right] F_{p-1,1} \right]$$

where:

$$H_p(x_i) = \frac{\partial}{\partial x_i} \left( J_{a^{(p)}}\left(z^{(p)}\right) J_{z^{(p)}}\left(a^{(p-1)}\right) \right)$$

and

$$F_{k,p} = \prod_{l=k}^{p} \left[ J_{a^{(l)}}\left(z^{(l)}\right) J_{z^{(l)}}\left(a^{(l-1)}\right) \right]$$

The following equations present the fully expanded chain rule corresponding to the Jacobian representation introduced above.

First derivative:

$$\frac{\partial y}{\partial x_i} = \frac{\partial y}{\partial z_m^{(k)}} \sum_n \frac{\partial z_m^{(k)}}{\partial a_n^{(k\text{-}1)}} \frac{\partial a_n^{(k\text{-}1)}}{\partial z_n^{(k\text{-}1)}} \cdots \sum_p \frac{\partial z_r^{(3)}}{\partial a_p^{(2)}} \frac{\partial a_p^{(2)}}{\partial z_p^{(2)}} \sum_q \frac{\partial z_p^{(2)}}{\partial a_q^{(1)}} \frac{\partial a_q^{(1)}}{\partial z_q^{(1)}} \frac{\partial z_q^{(1)}}{\partial x_i}$$

Second derivative:

$$\frac{\partial^2 y}{\partial x_i^2} = \left[ \frac{\partial y}{\partial z_m^{(k)}} \right]' \sum_n \frac{\partial z_m^{(k)}}{\partial a_n^{(k\text{-}1)}} \frac{\partial a_n^{(k\text{-}1)}}{\partial z_n^{(k\text{-}1)}} \cdots \sum_p \frac{\partial z_r^{(3)}}{\partial a_p^{(2)}} \frac{\partial a_p^{(2)}}{\partial z_p^{(2)}} \sum_q \frac{\partial z_p^{(2)}}{\partial a_q^{(1)}} \frac{\partial a_q^{(1)}}{\partial z_q^{(1)}} \frac{\partial z_q^{(1)}}{\partial x_i}$$
$$+ \frac{\partial y}{\partial z_m^{(k)}} \left[ \sum_n \frac{\partial z_m^{(k)}}{\partial a_n^{(k\text{-}1)}} \frac{\partial a_n^{(k\text{-}1)}}{\partial z_n^{(k\text{-}1)}} \cdots \sum_p \frac{\partial z_r^{(3)}}{\partial a_p^{(2)}} \frac{\partial a_p^{(2)}}{\partial z_p^{(2)}} \sum_q \frac{\partial z_p^{(2)}}{\partial a_q^{(1)}} \frac{\partial a_q^{(1)}}{\partial z_q^{(1)}} \frac{\partial z_q^{(1)}}{\partial x_i} \right]'$$

To distinguish between neurons and layers:

- **Subscripts** index neurons within a layer:
    - $i$ for input features
    - $q$ for neurons in the first hidden layer (layer 2)
    - $p$ for neurons in the second hidden layer (layer 3)

- $m$ for neurons in the output layer (layer $k$)

- **Superscripts** indicate the layer number:
  - Layer 1 → input features
  - Layer 2 → first hidden layer
  - Layer $k$ → final/output layer

**S.6: Visualizing the Impact of the Hyperparameter Beta on the PINN**

The influence of the hyperparameter beta (β) on model performance ($R^2$ and MSE) was assessed using five randomly generated hyperparameter sets, each derived from the defined search space bounds. The beta values tested were 1, 10, 100, 1000, and 10000; among these, 1, 10, and 100 were previously employed in Jiang et al. [16], while 1000 and 10000 were included to confirm that arbitrarily increasing β does not necessarily enhance accuracy. For every combination of hyperparameter set and beta value, models were trained over multiple random states and data splits. Importantly, no hyperparameter was explicitly tuned for any given β; this prevents results from being biased toward any β, allowing the observed differences in $R^2$ and MSE to be attributed solely to changes in β rather than to simultaneous adjustments in other model settings. A lower $R^2$ (0.2-0.5) is expected for both the test and training data, as none of the hyperparameters have been tuned.

The results are illustrated in Fig. 7, which shows the mean $R^2$ and MSE across all datasets, with error bars representing their standard deviations. It was observed that $\beta = 100$ yields the highest average accuracy and the least variability among the other values. This behaviour can be explained by the structure of the PINN loss function, which is the sum of the neural network's data-fitting loss and the PDE-residual loss. From Equations (3) and (7) in the main script:

$$G(x)=\frac{x}{1+e^{\beta|x|}} \tag{15}$$

$$Loss_{PDE} = L\left(0, G_{out}\left(\frac{\partial N}{\partial \epsilon}\right)\right) + L\left(0, G_{out}\left(\frac{\partial^2 N}{\partial \epsilon^2}\right)\right) + L\left(0, G_{out}\left(\frac{\partial N}{\partial T}\right)\right) + L\left(0, G_{out}\left(\frac{\partial N}{\partial d}\right)\right) \tag{16}$$

As β increases, the output of G(x) approaches zero, causing the PDE-residual loss to diminish. This weakens the physics constraints in the PINN, reducing its ability to generalize according to the governing equations and thereby lowering accuracy. Conversely, when β is too small, the PDE-residual loss becomes dominant, leading the model to overemphasise the physics constraints and neglect noisy or imperfect training data, resulting in higher error and reduced accuracy. The value β = 100 appears to strike an optimal balance between data fidelity and physical consistency, resulting in superior, more stable performance.

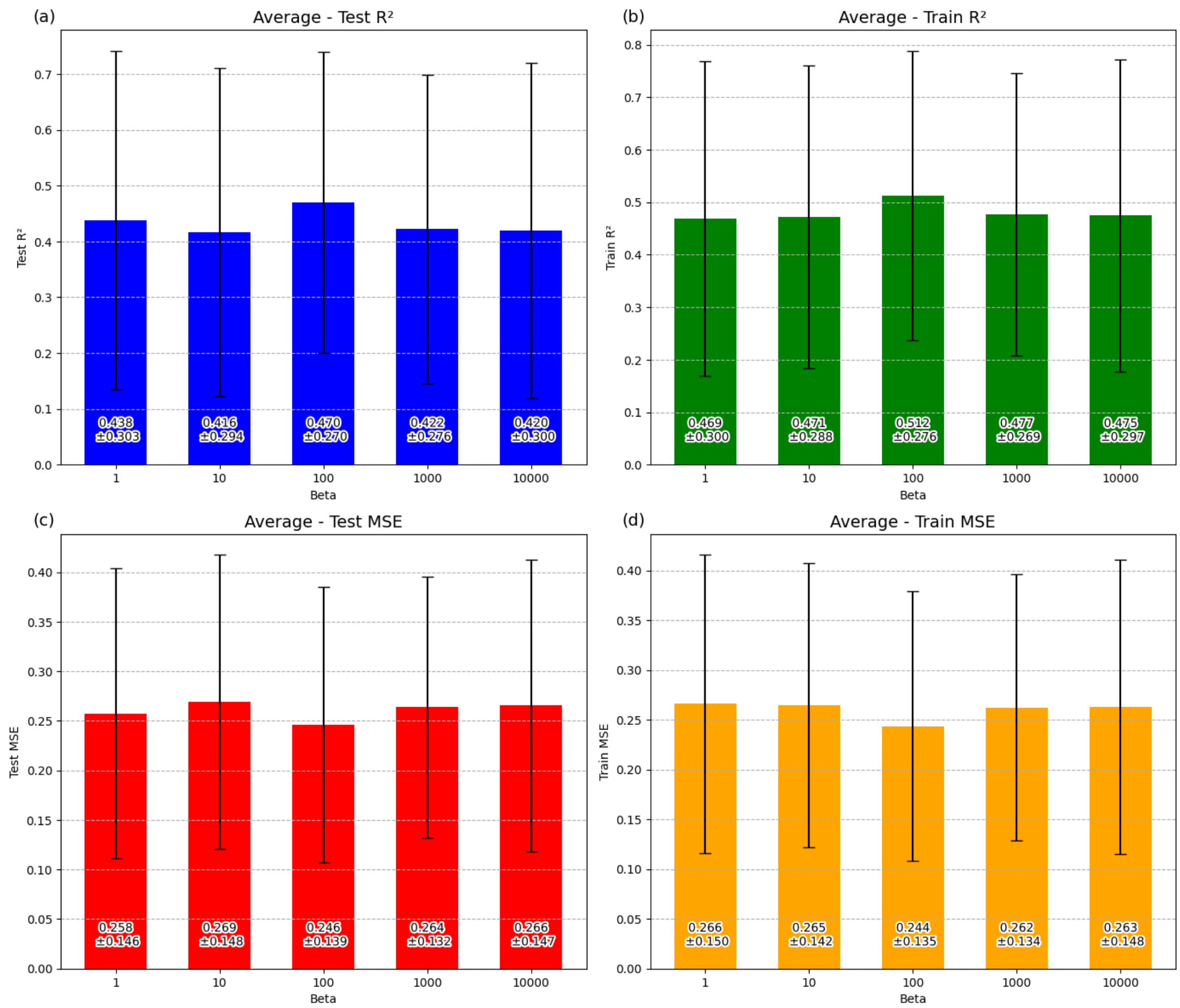

Fig. S.6: Bar plots showing the average values with standard deviation error bars for (a) Test $R^2$, (b) Train $R^2$, (c) Test MSE, and (d) Train MSE, obtained by varying the hyperparameter β across multiple random states, data splits, and randomized hyperparameter sets. The β values considered are 1, 10, 100, 1000, and 10000. The text within each bar indicates the average value and its standard deviation. Notably, Beta = 100 yields the highest average performance with the lowest variability.

## S.7: Hyperparameter Tuning - Bayesian Search Optimization

Hyperparameter tuning is the process of identifying the optimal combination of hyperparameters that yields the best performance for a given machine learning model on a specific dataset. Choosing the correct hyperparameters is crucial, as they govern the model's capacity, learning dynamics, and overall generalization ability.

Traditional techniques like **Grid Search** and **Randomized Search** evaluate multiple hyperparameter combinations independently and without leveraging information from previous evaluations:

- Grid Search exhaustively explores all possible combinations within a predefined grid.
- Randomized Search samples a fixed number of random combinations from the hyperparameter space, offering faster results with potentially similar performance.

In contrast, **Bayesian Optimization** [17], introduces a more intelligent and efficient strategy by using prior evaluations to guide future exploration. It builds a **probabilistic surrogate model** (commonly a Gaussian Process Regression (GPR) or a Tree-structured Parzen Estimator) to approximate the true objective function, which maps hyperparameter sets to model performance.

This process iteratively evaluates the model's performance across different hyperparameter configurations, updates the surrogate model based on observed results, and selects the next set of hyperparameters to assess using the acquisition function. This iterative process continues until a predefined stopping criterion is met, such as reaching a maximum number of iterations or achieving a satisfactory level of performance.

**Cross-validation** is often employed to obtain reliable estimates of the model's performance for each hyperparameter configuration. Cross-validation involves:

- Partitioning the data into training and validation data,
- Training the model on the training data and
- Evaluating its performance on the validation data.

This process is repeated with different training-validation splits to mitigate overfitting and assess the model's robustness.

As a result, Bayesian search is typically faster and more robust than exhaustive or random search methods, especially when evaluating each model is computationally expensive.

Table S.3: Hyperparameter Search Space for the RF model.

| RF | Hyperparameter Search Space |
|---|---|
| Min_samples_leaf | 1 - 5 |
| Min_ samples_split | 2 - 5 |
| N_estimators | 50-150 |

Table S.4: Hyperparameter Search Space for the GB model.

| GBoost | Hyperparameter Search Space |
|---|---|
| lr | 0.01-0.0001 |
| Max depth | 3-8 |
| Min_samples_leaf | 1-5 |
| N_estimators | 50-150 |

Table S.5: Hyperparameter Search Space for the XGBoost model.

| XGBoost | Hyperparameter Search Space |
|---|---|
| Max depth | 3-8 |
| N_estimators | 50-150 |
| lr | 0.0001-0.01 |
| Gamma | 0-0.0001 |
| Lambda | 1-3 |
| Min_child_weight | 3-8 |

Table S.6: Hyperparameter Search Space for the NN model.

| NN | Hyperparameter Search Space |
|---|---|
| Hidden dim 1 | 100-120 |
| Hidden dim 2 | 60-80 |
| Hidden dim 3 | 20-40 |
| Kappa | 0.1-0.9 |
| Batch size | 20-50 |
| Number of Epochs | 900-1000 |
| Weight Decay | 0.001-0.1 |

Table S.7: Hyperparameter Search Space for the PINN model.

| PINN | Hyperparameter Search Space |
|---|---|
| Hidden dim 1 | 100-120 |
| Hidden dim 2 | 60-80 |
| Hidden dim 3 | 20-40 |
| Kappa | 0.1-0.9 |
| Batch size | 20-50 |
| Number of Epochs | 900-1000 |
| Weight Decay | 0.001-0.1 |
| Omega | 0.01-0.99 |

## S.8: Accuracies obtained from all the data splits of all the employed models

Table S.8: Utilised Best Hyperparameters for different data splits for NN and PINN, obtained by Bayesian Search. The same hyperparameters are used for both NN and PINN models, except for Best Omega.

| Random State | Data Split | Best Hidden Dim 1 | Best Hidden Dim 2 | Best Hidden Dim 3 | Kappa | Batch Size | Number of Epochs | Weight Decay |
|---|---|---|---|---|---|---|---|---|
| 1 | 80:20 | 100 | 60 | 34 | 0.9 | 44 | 900 | 0.004436 |
| 1 | 70:30 | 116 | 75 | 31 | 0.446 | 40 | 965 | 0.001795 |
| 2 | 80:20 | 118 | 71 | 34 | 0.891 | 30 | 932 | 0.00247 |
| 2 | 70:30 | 120 | 78 | 40 | 0.723 | 37 | 1000 | 0.002501 |
| 3 | 80:20 | 100 | 80 | 37 | 0.9 | 20 | 1000 | 0.001 |
| 3 | 70:30 | 100 | 61 | 20 | 0.9 | 20 | 900 | 0.002558 |
| 4 | 80:20 | 115 | 64 | 20 | 0.278 | 21 | 966 | 0.001164 |
| 4 | 70:30 | 100 | 65 | 30 | 0.1 | 30 | 1000 | 0.001 |

Table S.9: Accuracies ($R^2$) obtained for testing and training data for all the models used in benchmarking.

| Random State | Data Split | Test NN $R^2$ | Train NN $R^2$ | Test RF $R^2$ | Train RF $R^2$ | Test GB $R^2$ | Train GB $R^2$ | Test XGB $R^2$ | Train XGB $R^2$ |
|---|---|---|---|---|---|---|---|---|---|
| 1 | 80:20 | 0.784 | 0.915 | 0.841 | 0.976 | 0.799 | 0.949 | 0.833 | 0.997 |
| 1 | 70:30 | 0.789 | 0.914 | 0.840 | 0.974 | 0.827 | 0.949 | 0.849 | 0.998 |
| 2 | 80:20 | 0.765 | 0.905 | 0.797 | 0.975 | 0.790 | 0.946 | 0.771 | 0.996 |
| 2 | 70:30 | 0.753 | 0.858 | 0.806 | 0.972 | 0.820 | 0.946 | 0.832 | 0.996 |
| 3 | 80:20 | 0.872 | 0.921 | 0.895 | 0.975 | 0.886 | 0.935 | 0.901 | 0.996 |
| 3 | 70:30 | 0.836 | 0.900 | 0.872 | 0.975 | 0.877 | 0.940 | 0.867 | 0.996 |
| 4 | 80:20 | 0.849 | 0.874 | 0.866 | 0.973 | 0.851 | 0.939 | 0.847 | 0.995 |
| 4 | 70:30 | 0.815 | 0.921 | 0.826 | 0.968 | 0.815 | 0.941 | 0.838 | 0.995 |

Table S.10: Accuracies (MSE) obtained for testing and training data for all the models used in benchmarking.

| Rando m State | Data Split | Test NN MSE | Train NN MSE | Test RF MSE | Train RF MSE | Test GB MSE | Train GB MSE | Test XGB MSE | Train XGB MSE |
|---|---|---|---|---|---|---|---|---|---|
| 1 | 80:20 | 0.0958 | 0.0430 | 0.0706 | 0.0119 | 0.0892 | 0.0259 | 0.0742 | 0.0014 |
| 1 | 70:30 | 0.1085 | 0.0415 | 0.0824 | 0.0127 | 0.0888 | 0.0249 | 0.0775 | 0.0012 |
| 2 | 80:20 | 0.1145 | 0.0466 | 0.0990 | 0.0124 | 0.1025 | 0.0265 | 0.1115 | 0.0021 |
| 2 | 70:30 | 0.1248 | 0.0687 | 0.0978 | 0.0136 | 0.0910 | 0.0262 | 0.0848 | 0.0019 |
| 3 | 80:20 | 0.0439 | 0.0417 | 0.0362 | 0.0134 | 0.0392 | 0.0344 | 0.0340 | 0.0019 |
| 3 | 70:30 | 0.0580 | 0.0549 | 0.0451 | 0.0137 | 0.0436 | 0.0328 | 0.0470 | 0.0020 |
| 4 | 80:20 | 0.0739 | 0.0624 | 0.0656 | 0.0132 | 0.0726 | 0.0304 | 0.0748 | 0.0027 |
| 4 | 70:30 | 0.0958 | 0.0383 | 0.0901 | 0.0156 | 0.0957 | 0.0284 | 0.0839 | 0.0022 |

Table S.11: Accuracies ($R^2$ and MSE) obtained for testing and training data for PINN alongside the Best omega (obtained through Bayesian Search) used in its calculation.

| Random State | Data Split | Best omega in (0.01-0.99) | Test PINN $R^2$ | Train PINN $R^2$ | Test PINN MSE | Train PINN MSE |
|---|---|---|---|---|---|---|
| 1 | 80:20 | 0.05 | 0.820 | 0.922 | 0.0800 | 0.0393 |
| 1 | 70:30 | 0.23 | 0.847 | 0.905 | 0.0784 | 0.0462 |
| 2 | 80:20 | 0.09 | 0.820 | 0.913 | 0.0879 | 0.0428 |
| 2 | 70:30 | 0.19 | 0.832 | 0.920 | 0.0849 | 0.0386 |
| 3 | 80:20 | 0.11 | 0.863 | 0.908 | 0.0470 | 0.0490 |
| 3 | 70:30 | 0.21 | 0.860 | 0.901 | 0.0495 | 0.0545 |
| 4 | 80:20 | 0.02 | 0.880 | 0.889 | 0.0588 | 0.0548 |
| 4 | 70:30 | 0.1 | 0.851 | 0.931 | 0.0769 | 0.0335 |

## S.9: Training Convergence of Loss Components in the PINN Framework

The convergence behaviour of the loss functions corresponding to the three components i.e., data-driven loss (NN), physics-informed loss (PDE) and the combined total loss of the PINN during the training process are analyzed. Below are the figures showcasing the convergence curves.

Although the physics-informed loss converges to values on the order of $10^{-6}$, this behavior is intrinsic to the sigmoid formulation defined in Eqs. (11) – (12) in the article, together with the Huber loss.

Once the PINN satisfies the imposed monotonicity and consistency conditions, the sigmoid rapidly saturates, and the Huber loss suppresses remaining residuals, leading to a small steady-state PDE loss. Therefore, this behaviour indicates successful enforcement of the physical constraints rather than their relaxation during optimisation. The PDE-loss convergence plot shows a rapid initial decrease followed by early stabilisation, indicating that the network learns the problem's physical structure early in training. After this point, subsequent optimization primarily refines the data fit while remaining close to the physically admissible manifold. This is also reflected in the total PINN loss curve, where the early reduction is driven jointly by both components. In contrast, the data-driven term under already-satisfied physics dominates later improvements. The PINN model shows faster convergence and lower stable loss than the NN.

To further demonstrate that the model does not degenerate into a purely data-driven neural network, comparative studies with conventional ML models alongside the utilised NN are shown in Section 3.1 of the article, where the proposed physics-informed approach consistently exhibited improved prediction accuracy and robustness. The concerned portion is shown below.

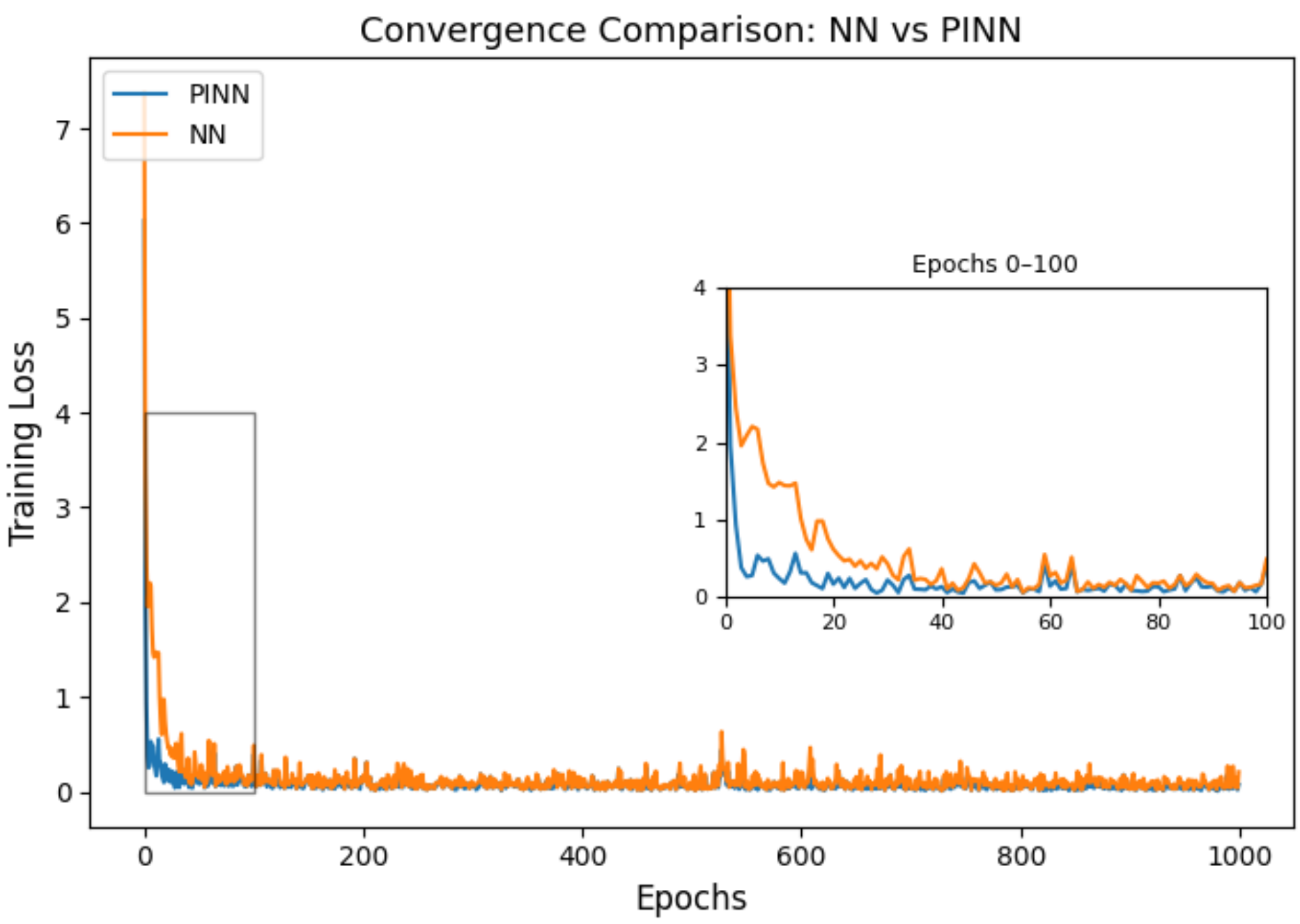

Fig. S.7: Convergence plots obtained from purely data-driven loss (NN) and the sum of data-driven and physics-informed PDE-loss (PINN). The inset shows a magnified view of the early training stage (epochs 0–200, loss 0–4). PINN converges faster and achieves higher accuracy than NN.

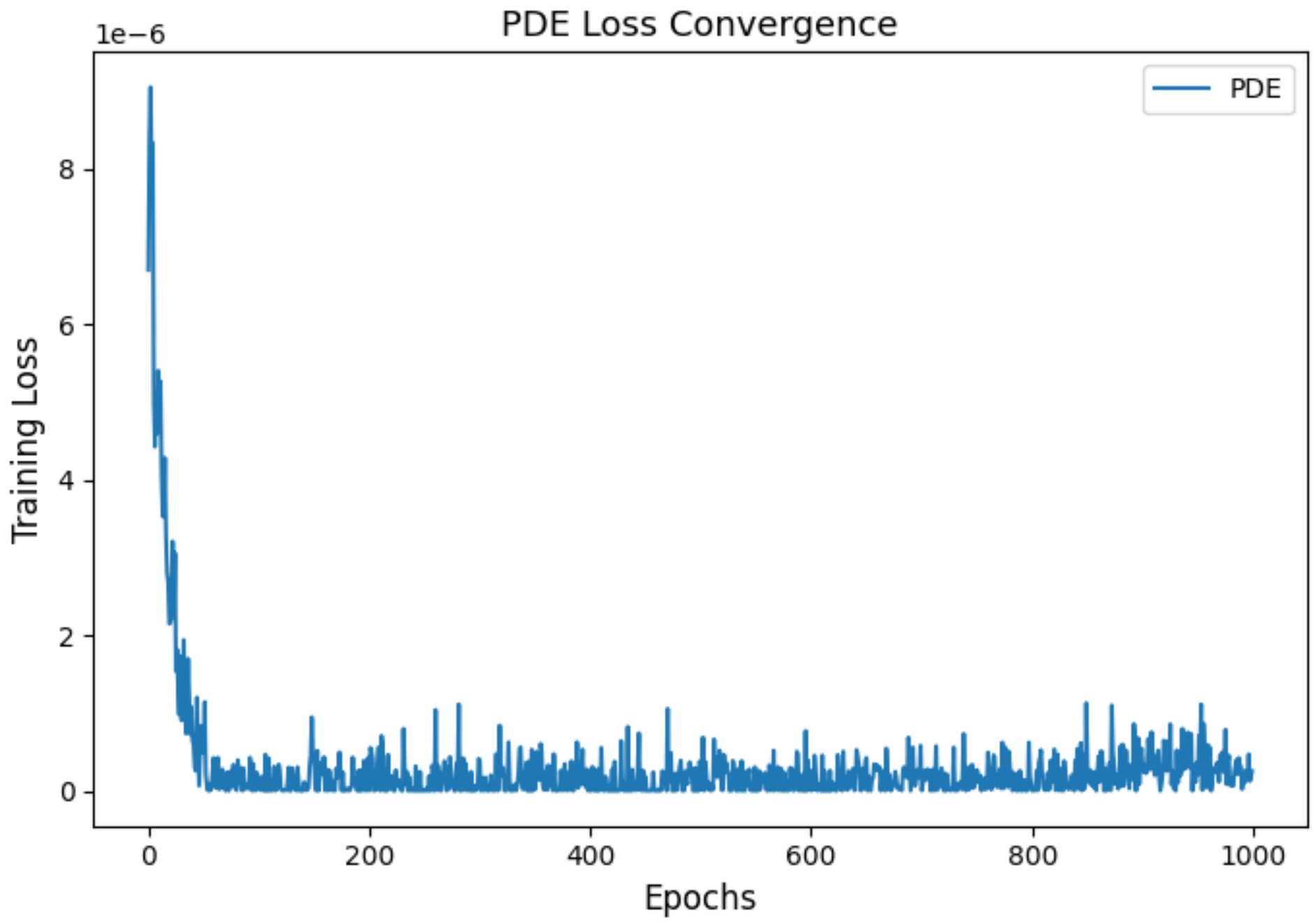

Fig. S.8: Convergence plot obtained from pure PDE-loss. While the loss values show a rapid initial decrease, they then stabilise early, demonstrating that the network learns the problem's physical structure at an early stage of training.

### S.10: Input-Feature Selection for Dimensionality Reduction

Feature selection was performed in a two-stage process. First, highly correlated features were identified using a correlation threshold of 0.7 to minimize redundancy. For each strongly correlated pair, the feature with the lower variance was removed. The corresponding correlation-matrix maps are presented in Fig. S.9. This procedure reduced the original feature set from 50 to 23, removing 27 features.

Following feature reduction, a slight decrease in average predictive accuracy was observed for all five models. As shown in Fig. S.10, RF, GBoost, and XGBoost continued to exhibit higher average accuracy than the PINN, while the NN showed the poorest performance. Notably, despite this reduction, the PINN maintained a high predictive capability, with the coefficient of determination ($R^2$) remaining above 0.8. In addition, the PINN exhibited the lowest variance among all models, consistent with its behaviour when trained on the complete feature set.

This sensitivity to feature removal suggests that the PINN relies on correlated, physics-relevant descriptors to enforce physical consistency. Removing such features results in a loss of information that the PINN, unlike other models, is less able to compensate for. This behaviour further indicates that the PINN does not overfit to a small subset of highly ranked features identified by SHAP. Instead, it leverages a broader feature space to achieve stable, physically meaningful predictions.

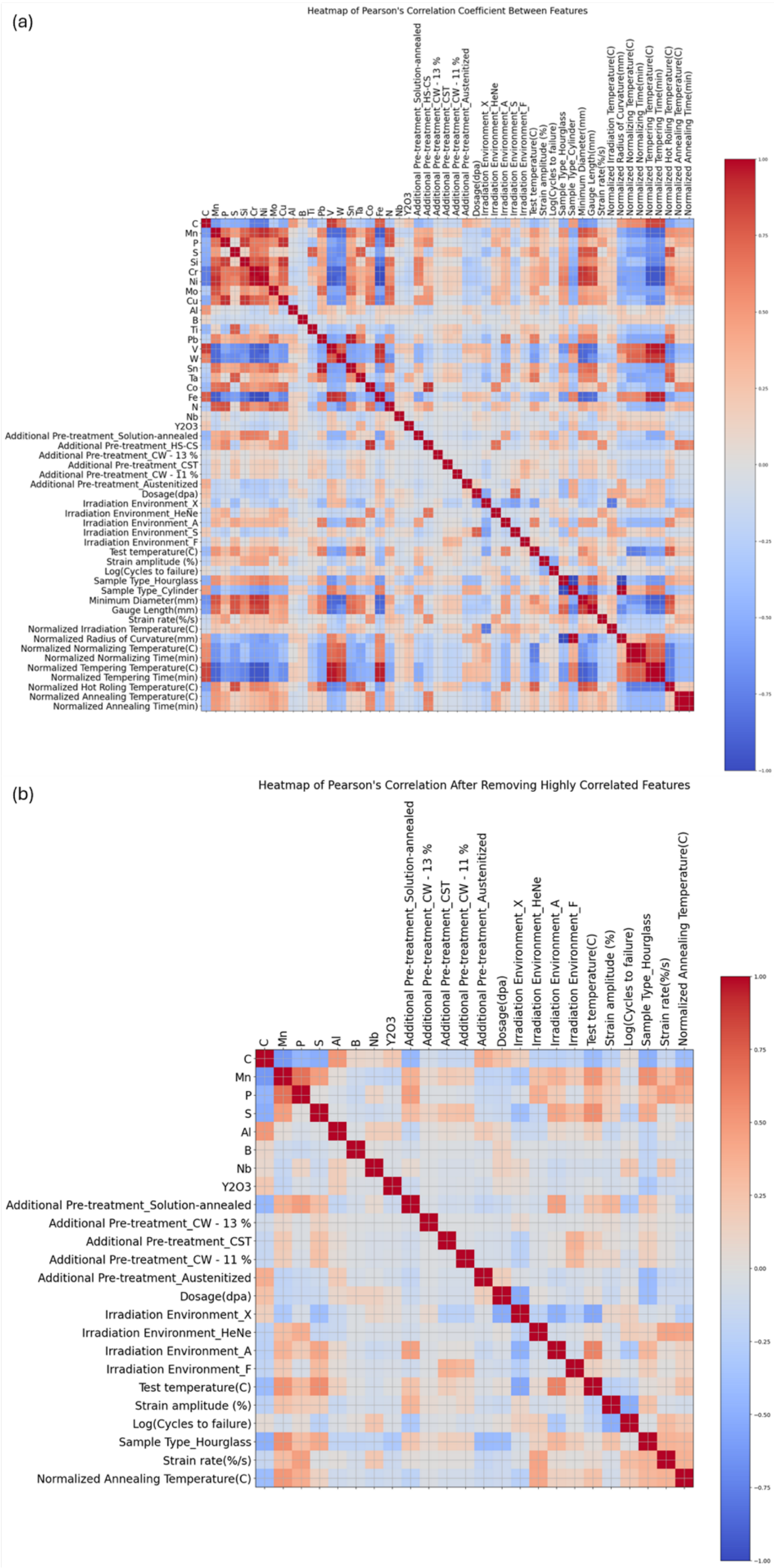

Fig. S.9: Visualization of Pearson's correlation coefficients considering (a) all the input features, (b) Input features obtained after the elimination of the features with lower variance from each highly correlated pair.

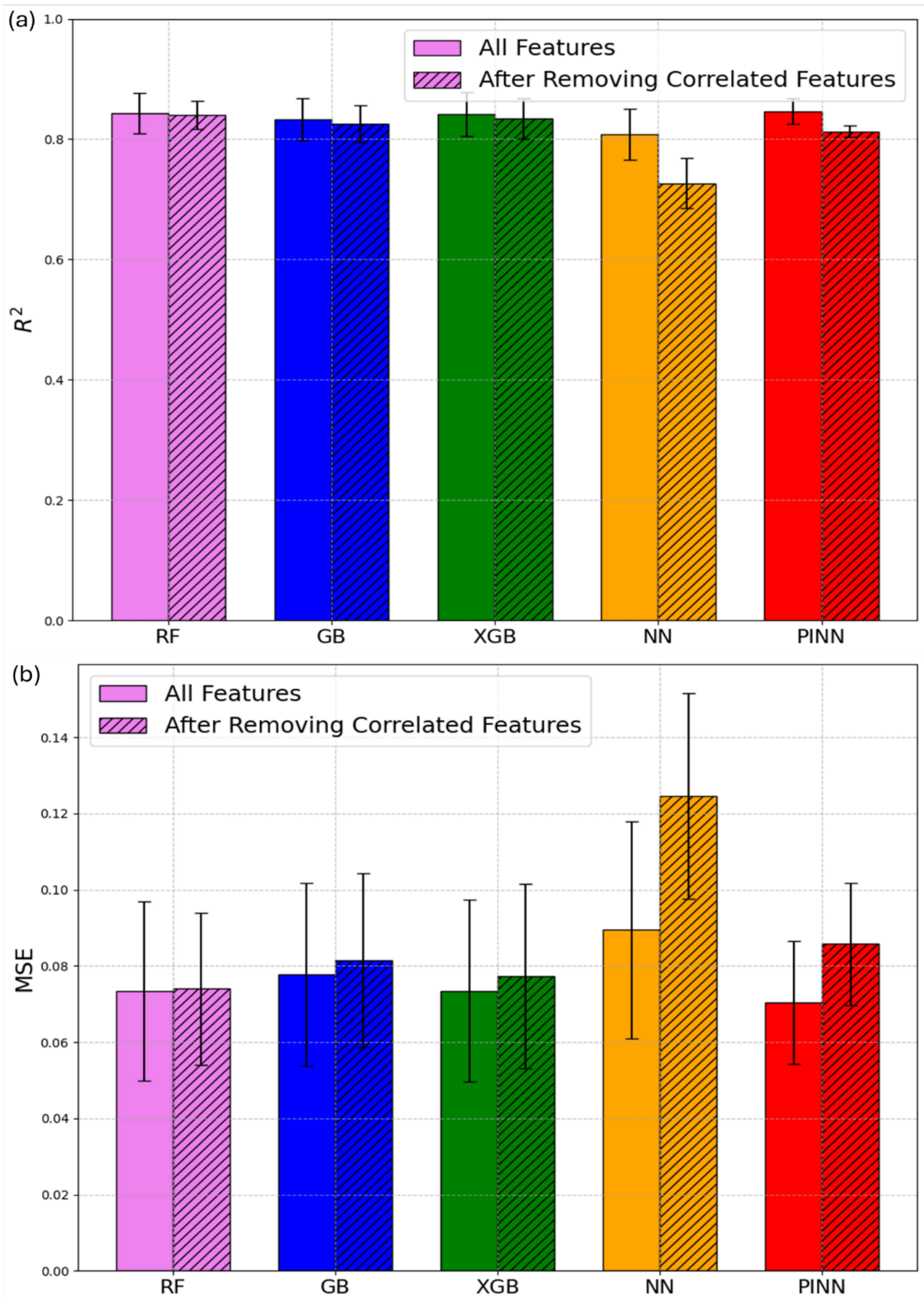

Fig. S.10: The average (a) $R^2$ and (b) MSE values for 5 predictive models before and after the removal of highly correlated input features: RF, GBoost, XGBoost, NN, and PINN tested across the combinations of all random states (1 to 4) and data-splits (70:30 and 80:20), shown with error bars representing the variance.